\documentclass[letterpaper, 10 pt, conference]{ieeeconf}  % Comment this line out if you need a4paper

\IEEEoverridecommandlockouts                              % This command is only needed if 
\usepackage{amsmath,amsfonts,graphicx}
\usepackage{amssymb}  % assumes amsmath package 
\usepackage{graphics} % for pdf, bitmapped graphics files
\usepackage{stackengine}
\usepackage{graphicx} 
\newcommand*{\Scale}[2][4]{\scalebox{#1}{$#2$}}%
\usepackage{times} % assumes new font selection scheme installed
\usepackage{esvect}
\usepackage{comment}
\usepackage{algorithmicx}
\usepackage{algpseudocode}
\usepackage{graphics}
\usepackage[ruled,vlined]{algorithm2e}
\usepackage{graphicx,wrapfig,lipsum}
\usepackage{xcolor}
\usepackage{mathtools}
\usepackage{textgreek}
\usepackage{bm}
\usepackage{courierten}

\usepackage{url}

\usepackage{color}
\DeclareMathAlphabet{\mathcal}{OMS}{cmsy}{m}{n}
\DeclareMathAlphabet{\mathbcal}{OMS}{cmsy}{b}{n}

\usepackage{cite}
\usepackage{dblfloatfix}
\definecolor{Blue}{rgb}{0,0,0}
\definecolor{Black}{rgb}{1,1,1}
\usepackage[most]{tcolorbox}
\usepackage{array}

\usepackage{todonotes}
\usepackage{multirow}
\usepackage{subcaption}
\usepackage{comment}

\usepackage[most]{tcolorbox}
\definecolor{gold}{RGB}{255,215,0}
\definecolor{bright_yellow}{RGB}{255,255,0}
\definecolor{teal}{RGB}{61,126,182}
\definecolor{pale_green}{RGB}{27,197,148}
\definecolor{violet}{RGB}{188,126,193}
\definecolor{yellow_gray}{RGB}{163,163,153}

\title{\LARGE \bf
Active and Interactive Mapping with Dynamic Gaussian Process Implicit Surfaces for Mobile Manipulators}

\author{Liyang Liu$^1$, Simon Fryc$^1$, Lan Wu$^1$, Thanh Vu$^1$, Gavin Paul$^1$ and Teresa Vidal-Calleja$^1$
\thanks{$^1$All authors are with the Center for Autonomous Systems (CAS), University of Technology Sydney, Australia.
	 {\tt \small  \{Liyang.Liu, Simon.Fryc, Lan.Wu-1, Thanh.Vu, Gavin.Paul-1, Teresa.VidalCalleja\}@uts.edu.au}
  }
}

\begin{document}

\maketitle
\thispagestyle{empty}
\pagestyle{empty}

%%%%%%%%%%%%%%%%%%%%%%%%%%%%%%%%%%%%%%%%%%%%%%%%%%%%%%%%%%%%%%%%%%%%%%%%%%%%%%%%
\begin{abstract}
\textcolor{Blue}{In this letter, we present an interactive probabilistic mapping framework for a mobile manipulator picking objects from a pile. The aim is to map the scene, actively decide where to go next and which object to pick, make changes to the scene by picking the chosen object, and then map these changes alongside. }The proposed framework uses a novel dynamic Gaussian Process (GP) Implicit Surface method to incrementally build and update the scene map that reflects environment changes. Actively the framework provides the next-best-view, balancing the need for picking object reachability with map information gain (IG). To enforce a priority of visiting boundary segments over unknown regions, the IG formulation includes an uncertainty gradient-based frontier score by exploiting the GP kernel derivative.
This leads to an efficient strategy that addresses the often conflicting requirement of unknown environment exploration and object picking exploitation given a limited execution horizon. We demonstrate the effectiveness of our framework with software simulation and real-life experiments. 
\end{abstract}

%%%%%%%%%%%%%%%%%%%%%%%%%%%%%%%%%%%%%%%%%%%%%%%%%%%%%%%%%%%%%%%%%%%%%%%%%%%%%%%%
\section{\textcolor{Blue}{INTRODUCTION}}
\textcolor{Blue}{Recent years have seen great progress in autonomous mobile manipulation applications. Typical examples include bin-picking for construction sites \cite{simon-acra}, automated public space sanitisation \cite{xdbot} and logistic and warehousing \cite{warehouse}. These applications call for interactive robotic systems that are able to explore the scene while mapping the changing environment and planning their motion and manipulation tasks without colliding with obstacles and with limited on-board resources.} 
%Manipulation tasks may differ on grasping, picking, placing, material deposition. The task considered here is in the form of object picking.

%Regardless of the task at hand, an autonomous robotic manipulator needs first to explore the environment, generate a map of it and update the map once the scene has changed.
\textcolor{Blue}{Scene exploration approaches often adopt an information-theoretic strategy that aims to choose the next action to maximise the Information Gain (IG) in an active mapping framework~\cite{masha-ipp,Lauri2015ActiveOR,fred-apple-picking}. %This has been studied for terrain monitoring~\cite{masha-ipp}, active object detection~\cite{Lauri2015ActiveOR}, fruit detection~\cite{fred-apple-picking} and many other applications. 
For mobile manipulator tasks, active mapping alone is insufficient. Combining mapping and picking in one phase is more effective and efficient than considering them separately. Thus, the task becomes an active and interactive mapping problem, \emph{i.e.} to select and pick the ``best" objects with mapping aiding the selection, and where the next movement expands the knowledge of the scene.} 

\begin{figure}[t]
	\centering
	\setlength\tabcolsep{1pt}	
	\begin{tabular}{cc}	
	\tcbox[arc=1pt,top=-1pt,left=-1pt,right=-1pt,bottom=-1pt,colback=white,colframe=blue!70!green]{\includegraphics[clip, trim=0cm 2cm 0cm 0cm,width=0.45\linewidth]{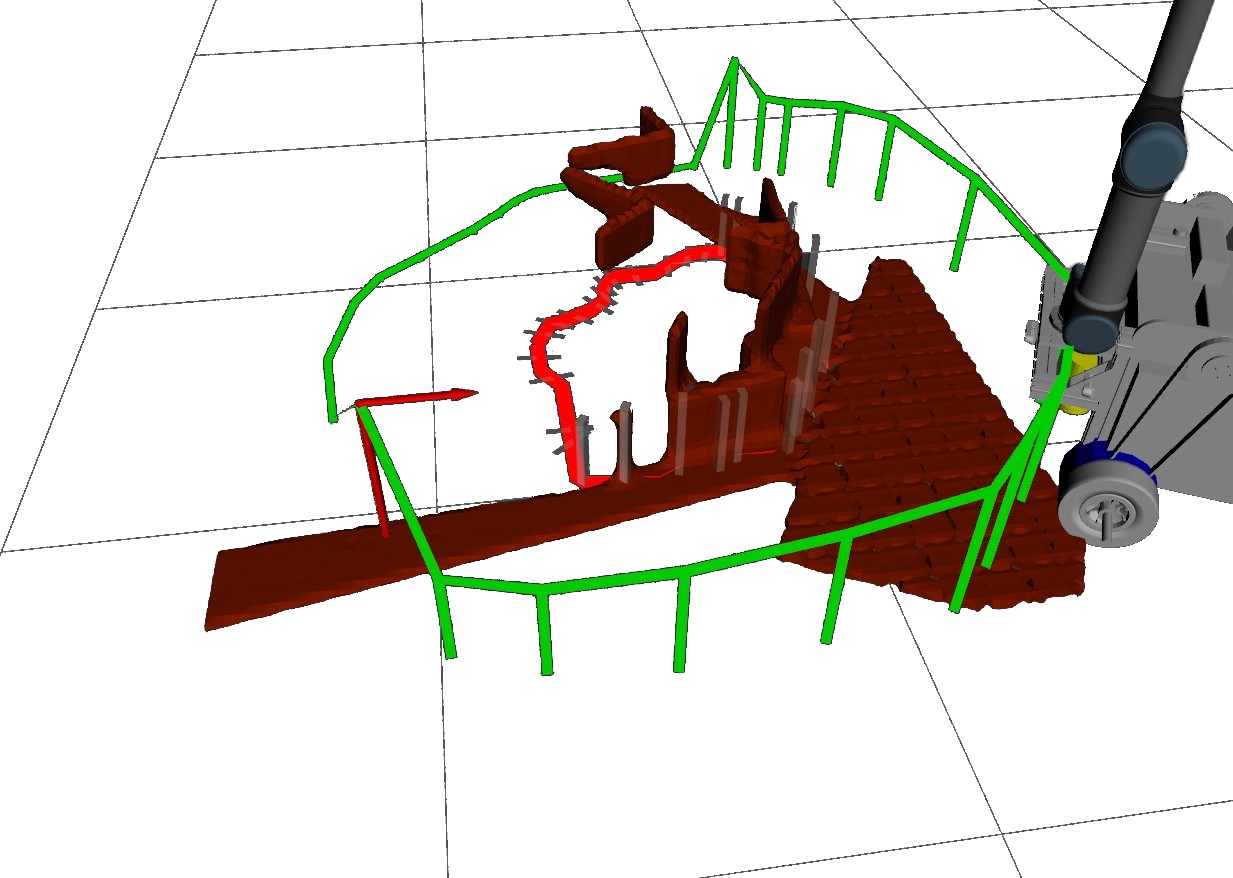}} &
	\tcbox[arc=1pt,top=-1pt,left=-1pt,right=-1pt,bottom=-1pt,colback=white,colframe=blue!70!green]{\includegraphics[clip, trim=0cm 2cm 0cm 0cm,width=0.45\linewidth]{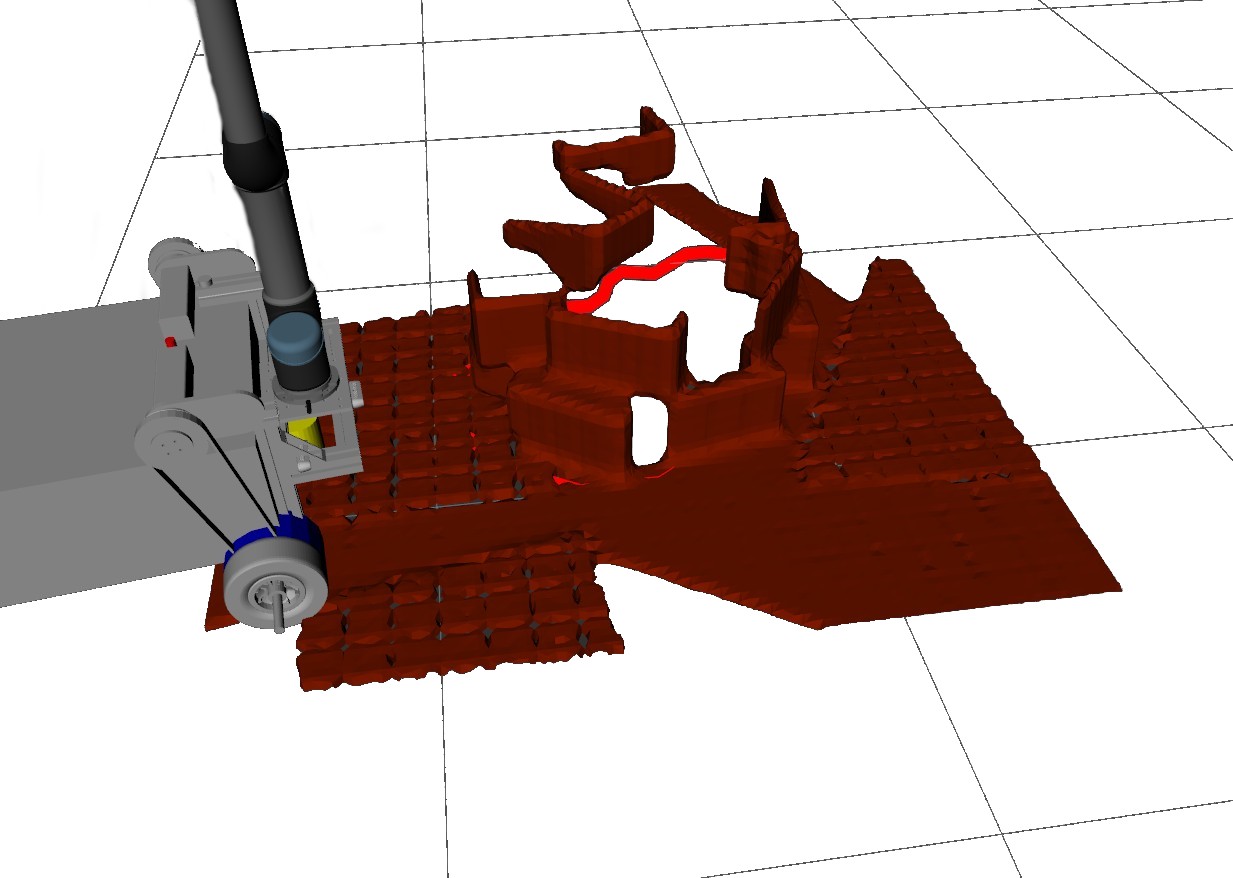}}  \\  [-7pt]
	\small{(\emph{a}) Initial map and NBV } 
& 	\small{(\emph{b}) Move to NBV and update map} \\
	\tcbox[arc=1pt,top=-1pt,left=-1pt,right=-1pt,bottom=-1pt,colback=white,colframe=blue!70!green]{\includegraphics[clip, trim=0cm 7cm 0cm 0cm,width=0.45\linewidth]{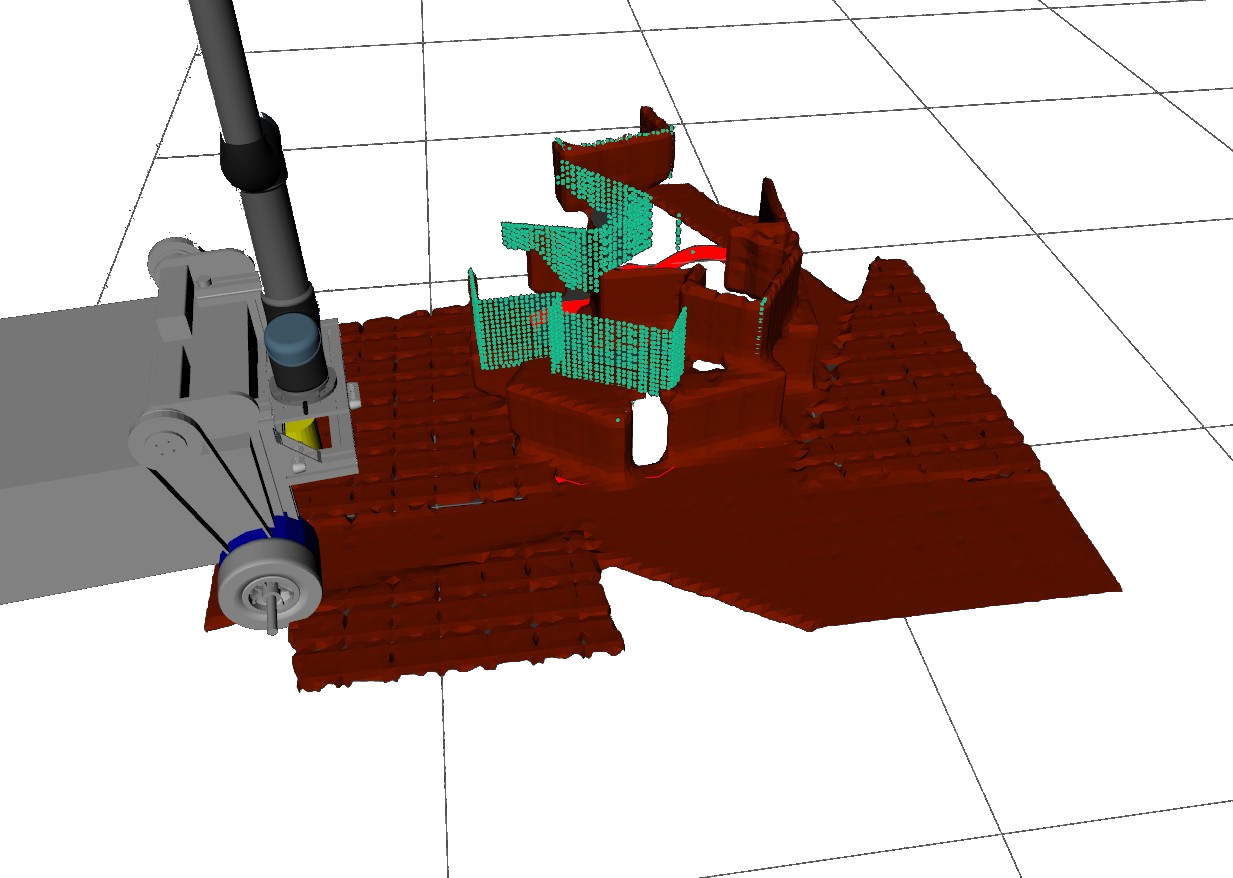}} &
	\tcbox[arc=1pt,top=-1pt,left=-1pt,right=-1pt,bottom=-1pt,colback=white,colframe=blue!70!green]{\includegraphics[clip, trim=0cm 7cm 0cm 0cm,width=0.45\linewidth]{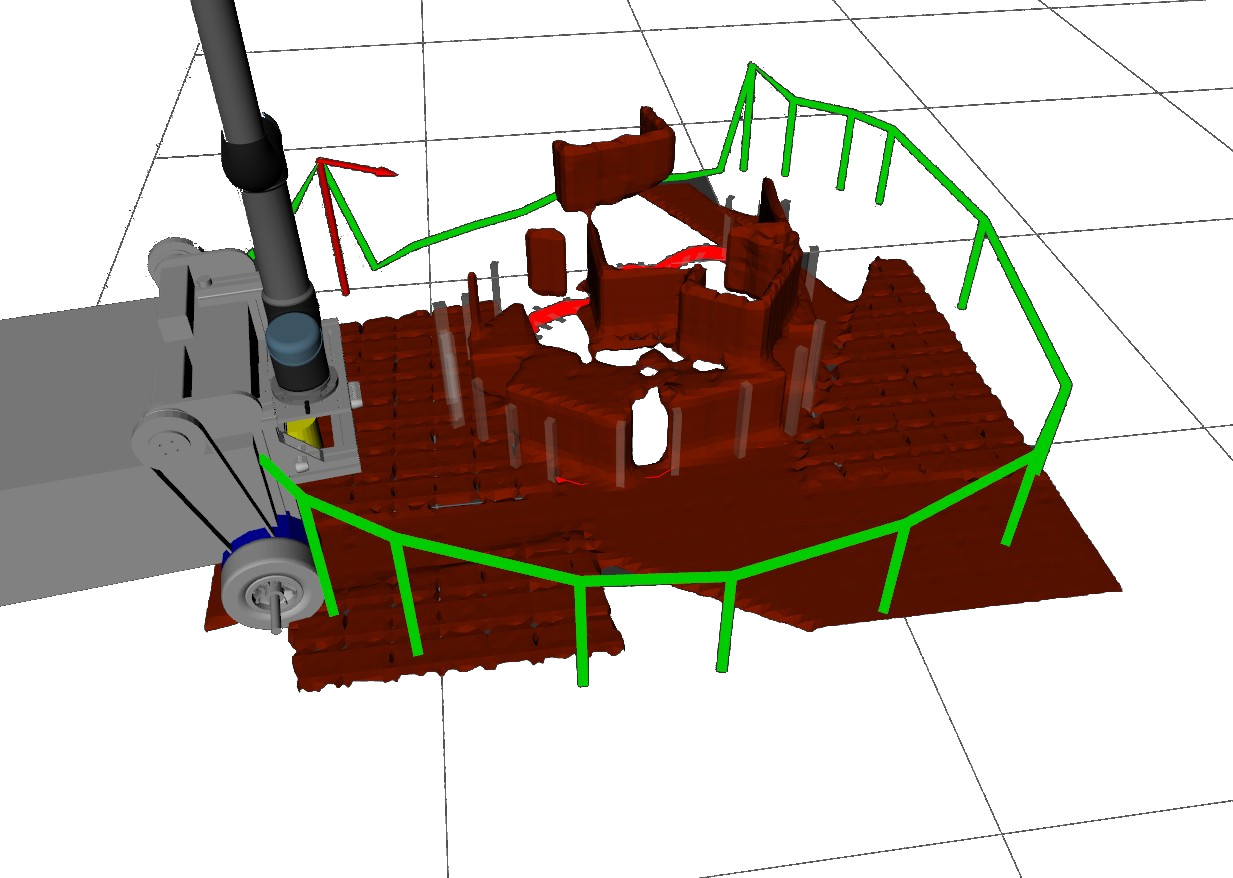}}  \\ [-7pt]
	\small{(\emph{c}) Remove object	} 
		& 	\small{(\emph{d}) Update map and NBV	} \\
	\end{tabular}			\caption{\small{The Active and Interactive Mapping cycle.
	%: build a GPIS from the initial scan, move to the NBV and update GPIS, remove objects, then repeat. 
	The projection of GPIS mesh on the ground (red line) defines potential exploration segments, green bars indicate segments information utilities. The NBV is a red bar + arrow. Dark green dots are GPIS training points from the removed object. }}
{\vspace{-10pt}}	
	\label{fig:gpis-demo}
\end{figure}

%Generally, an autonomous mobile manipulator needs to choose a

%

%\noindent \textbf{Contributions and Outline}\\
%This paper builds on previous work \cite{simon-acra} and presents a interactive probabilistic framework.

\textcolor{Blue}{
This paper presents a computationally efficient environment mapping framework for a mobile manipulator realised on low-budget hardware for practical bin-picking applications, such as is shown in Fig. \ref{fig:gpis-demo}. It aims to
address the problems of dynamic scene mapping, exploiting the map's frontier and checking manipulability to select the next best view for efficient mobile base placement and manipulator motion planning. Here, dynamic mapping refers to the accurate capture of environment changes as objects are discovered (scene exploration) and later removed from the scene (object picking) by a mobile manipulator}. 

%We propose a GPIS based mapping framework for exploration on low-budget hardware in a practical bin-picking application that
%
%Next-best-view (NBV) is a well-known problem in robotics that has to satisfy two main goals in the case of mobile manipulators: the chosen place needs to enable the robot to detect and pick objects, and with the on-board sensor's broad field-of-view (FOV), the robot should increase map coverage and improve map accuracy. Conflicts often arise when pursuing both goals. The new chosen place must be in a previously mapped and connected section of the map to allow the planning of a collision-free path, as well as contain as many manipulable objects as possible. The robot also needs to identify and inspect frontiers in order to expand its knowledge of the scene. This paper aims to tackle all these aspect by considering manipulator's manipulability and a flexible a probabilistic mapping approach.
%A NBV selection scheme must be carefully designed to be flexible and fair considering all factors involved for an optimum outcome. A naive approach based on rigid logic will not be adequate to cater for and interact with the continuously changing world. This paper aims to tackle all these constrains.
%
\textcolor{Blue}{Our proposed mapping approach is based on Gaussian Probabilistic Implicit Surfaces (GPIS)~\cite{GPIS}, which encode spatial correlation among input data and offer a probabilistic yet accurate map representation of the world in continuous form. We exploit GPIS to check if a mapped object resides in a robot's workspace. Also, its probabilistic formulation makes it amenable to active mapping based on IG to analytically search for the next-best-view (NBV) and optimal motion. }
%The map can be updated online using Bayesian fusion on scans from varying viewpoints \cite{bhoram-lee,LanRAL20}. One can query the GPIS for scene geometry and uncertainty at arbitrary resolution, visited or unexplored alike, then perform  trajectory planning with obstacle avoidance in mind using the dense map obtained. By virtue of its definition, GPIS offers a straight forward way to check if a mapped object resides in robots configuration space. On the other hand, the probabilistic formulation in GPIS makes it particularly amenable to active perception problems as the robot can use any IG based analytical methods to make flexible decision about the next optimal move.

\textcolor{Blue}{
The contribution of this work is threefold. First, a dynamic GPIS algorithm (Section~\ref{sec:dgp}) where each incoming
sensor scan forms an instantaneous GP hence defines a probabilistic tolerance layer for valid samples. 
%sensor scan is trained in an instantaneous GP that defines a probabilistic tolerance layer. 
This detects and discards previous GPIS training points taken from the scene, thus producing a resultant implicit surface accurately capturing environment changes (Fig.~\ref{fig:gpis-demo}\emph{(a)} and (\emph{d)}). 
Secondly, given the probabilistic map representation, we develop an analytical frontier score for each possible NBV candidate (Section~\ref{sec:sigma_gradient}) exploiting the GP kernel derivative, this greatly improves fairness and flexibility in NBV selection. 
Ultimately, the proposed framework is a viable means for a mobile manipulator to interact with an environment (Section~\ref{sec:interact}): planning base placement and generating arm's trajectory with obstacle avoidance capability, choosing the NBV that maximises object manipulability (Fig.~\ref{fig:gpis-demo}(\emph{b})). 
Results of the full on-line implementation of our framework are presented in Section~\ref{sec:evaluate}, demonstrating its performance in simulation and real experiments.
}

\section{\textcolor{Blue}{Related Work}}
%\textcolor{Blue}{
%\textcolor{\noindent Our work is related to dynamic mapping and NBV selection.}}
\noindent \textbf{\textcolor{Blue}{Dynamic mapping}}\\
%Our work is related to actively mapping and manipulating objects in a partially-observed scene with a mobile robot. 
%\textcolor{Blue}{Previous works on dynamic mapping have focused mainly in, 1) detecting and tracking of moving objects in static scene background~\cite{static-map-dynamic-object}~\cite{DynaSLAM}, 2) eliminating dynamic objects from the environment in order to build accurate static map~\cite{first-dynamic}, 3) dynamic traffic road modelling for safe path planning~\cite{hilbert-dynamic}. These categories do not address the problem of \textit{manipulating} objects from the partially observed scene.}
%None of these categories are relevant to the area of active mapping of an environment and removing objects from it. }
%Despite the focus on dynamic mapping, these works are less applicable to our problem of actively mapping a scene and removing the objects from it. 
%
%Such kind of map needs to be accurate and updated from time to time to account for removed objects. 
%
\textcolor{Blue}{The choice of mapping representation is critical for object picking tasks with mobile manipulators and needs the ability to update efficiently once objects are removed. The well-known RGB-D mapping system KinectFusion \cite{kinect-fusion} is an example for this purpose. The map is a volumetric type built in the form of Truncated Signed Distance Function (TSDF) data structure. Each voxel in space is time-averaged to smooth out the transient noise. 
As a side-effect the model handles dynamic scenes. Dramatic topological changes, however, will only appear after a significant delay.
%The high storage space and GPU hardware requirements prevent practical application in large-scale systems. 
Occupancy maps, specifically Octomap~\cite{octomap} are another discrete representation of the 3D world with a probability of occupancy for each voxel in space. A known limitation of the original form is that dynamic environments are not supported. An extension to Octomap was introduced in~\cite{simon-acra} for a bin-picking application. However, since structural correlations between nearby cells is not addressed, resolution and accuracy in Octomap are often compromised. One can not easily reason about object shape or perform detection~\cite{gpis-shape}. 
Gaussian Processes Occupancy Maps (GPOM), on the other hand, is a method of combining GP with occupancy mapping that is probabilistic and continuous but has a large computational complexity. 
%However, its definition requires a huge amount of training data and does not give sufficient shape information compared to GPIS. 
Hilbert Maps, first proposed in~\cite{ramos-2016} is a faster and simpler type of occupancy map using kernel approximation methods. Later in~\cite{hilbert-dynamic} it was extended to handle dynamic obstacles and model learning through regression, it finds application in  traffic environment modelling.} 
%It was noted in~\cite{bhoram-lee} that obtaining a precise surface boundary from Hilbert Maps can be problematic. } 
%We concern for its suitability to close-distance interactive applications where object picking and placing are critical. 

\textcolor{Blue}{Despite its advantages, the $O(n^3)$ computational complexity in GPIS~\cite{GPIS} has limited its applicability. 
%Kim \emph{et al.} suggested a divide and conquer strategy in~\cite{kim-overlap} to solve the scalability issue. Later
Lee~\emph{et al.}~\cite{bhoram-lee} provided an online GPIS implementation that stores data in small clusters for fast parallel processing and incrementally fuses successive scans from various viewpoints. 
%utilising parallel processing power of multi-core computers. 
Dynamic environments, however, are not tackled in this method -- once an object is mapped, it remains in the GPIS data even if it is physically removed from the workspace. Here, we consider the GP from a fundamental perspective as an immediate probabilistic conditioning tool on current training data~\cite{gp-book}. We will show that once a mechanism is devised to detect and filter out training points belonging to the removed object, the map can be instantly updated in the relevant local regions.}

\textcolor{Blue}{Other work in dynamic mapping has been focused on detecting and tracking moving objects in static scenes~\cite{static-map-dynamic-object}~\cite{DynaSLAM} or eliminating dynamic objects from the environment in order to build accurate static maps~\cite{first-dynamic}, which is less applicable to object picking with mobile manipulators.
}  
\\

\noindent \textbf{\textcolor{Blue}{Active Exploration}}\\
\textcolor{Blue}{In active exploration, it is desirable to formulate an IG metric to choose the optimal action. Occupancy maps and Octomaps, with their probabilistic occupancy representation, are often used in mapping and exploration applications~\cite{slam-occ,FSMI}. GP, with its continuous uncertainty formulation, has also been used recently for IG-based exploration works, including, entropy reduction~\cite{masha-ipp}, conditional entropy reduction~\cite{huang2019building} and mutual information maximisation~\cite{maani-gp}. In active shape modelling, \cite{Gandler-explore} proposed an efficient approach using constrained Variational Sparse GP and online kernel learning that preserves reconstruction accuracy.}

\textcolor{Blue}{Entropy gradient~\cite{entropy-gradient}, is another form of IG metric. 
The benefit is that it directly draws the robot towards the frontier between the explored and unexplored regions. Authors in \cite{entropy-gradient} devised an approximate gradient formulation for discretised 3D occupancy maps. Later, Jadidi~\emph{et al.}~\cite{maani-gp} generated continuous gradient frontier maps from a 2D GPOM variant, but did not provide an analytical expression due to its non-trivial map definition. They used the gradient map for NBV candidates identification but not in NBV utility calculations. In this paper, we derive the analytical gradient expression from 3D GPIS uncertainty for NBV candidates in 2D mobile manipulator exploration. The simple yet consistent frontier score, together with other IG metrics ensure NBV selection is well-balanced. }
\\

\noindent \textbf{\textcolor{Blue}{Interactivity}}\\
\textcolor{Blue}{Interactivity refers here to the robot manipulating objects in the environment. GPIS, with built-in surface normals~\cite{gpis-shape}, has been exploited for interactive applications in areas of shape detection, classification and validation. The GPIS-based framework presented in~\cite{active-grf-gpis} makes use of a robotic arm equipped with tactile sensors to explore and map the environment. Authors in \cite{huang2019building} show how to optimise for a reconstruction-aware manipulator trajectory that during a pick-and-place action maximally estimates the object's 3D geometry. Our work, in contrast, generates dense maps and use them for collision-free planning for manipulator trajectory. Further, these prior works assume that every object in the robot's workspace is reachable by the arm. Our framework instead selects the NBV that covers the maximum number of manipulable objects. 
\\
}
{\vspace{-5pt}}	
\section{System Overview and Problem Statement}
\label{sec:overview}
\begin{figure}[h]
{\vspace{-5pt}}	
\centering
\tcbox[top=-1pt,left=-1pt,right=-1pt,bottom=-1pt,colback=white,colframe=blue!50!white]{\includegraphics[clip, trim=0cm 0.4cm 0cm 0.2cm,height=4.3cm,width=0.9\linewidth]{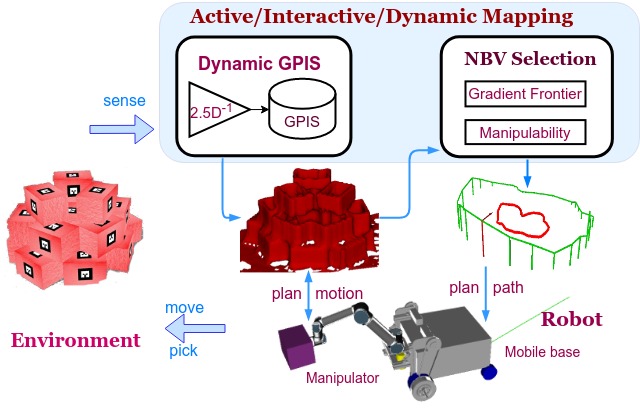}}
\caption{\small{Active and interactive mapping framework overview.}}
\label{fig:overview}
{\vspace{-5pt}}	
\end{figure}

%The proposed framework aims to provide an efficient object discovery and picking pipeline for a mobile manipulator operating in an un-mapped 3D scene comprised of objects stacked in a pile. 
The proposed framework aims to provide an efficient pipeline for a mobile manipulator to explore and interact with the environment. As illustrated in Fig.~\ref{fig:overview}, it consists of a GP-based dynamic mapping component that maps the changing world, and an NBV selection module that recommends the best action based on the map information. The robot relies on this framework to plan its arm trajectory and base paths to travel and make changes in the surrounding environment. 

%We now describe the framework's exploration sequence. 
Fig.~\ref{fig:gpis-demo} shows a robot's exploration and interactivity cycle. First, it inspects the scene to build the initial map of a multiple objects pile, and computes an NBV that will increase map information. Next, it moves to the recommended NBV and updates the map. Then, it picks up an object and updates the map again to synchronise with the changed world. Based on the updated map, it computes the new NBV and moves towards it, after which the inspection cycle repeats.

\textcolor{Blue}{We now define the notations in this paper:
a line segment $\mathbf{s}_i$ describes the robot's pose due to its finite width, $\mathbf{s}_i$ aligns with the base width-wise. The segment set  $\mathcal{S}=\{\mathbf{s}_i\}$ denotes candidates in NBV selection.
Superscript ${\{\cdot\}}^{[t]}$ indicates the item is time-dependent, and $\mathbf{s}_i^{[t]}$ is the $i$'th segment at $t$. The sensor pose $\mathbf{T}^{[t]} \in \mathrm{SE}(3)$ is assumed given at all times.
%\textcolor{Blue}{Throughout this paper we assume the depth sensor pose is given at all times}.
%$\{\cdot\}[s(i)^{[t]}]$ means  attribute $\{\cdot\}$ of the time dependent segment $\mathbf{s}(i)$ at time $t$.
%
$U^{[t]}(\, \mathbf{s}_i\, )$ denotes the information utility at time $t$ for the $i$'th segment. The selection goal is to identify the segment $i^{[t]}_\ast$ with the maximum utility to position the robot at $t$,
\begin{equation}
\begin{aligned}
    i^{[t]}_\ast &= \underset{i}{\text{argmax}}\, {U^{[t]}(\,\mathbf{s}_i^{[t]}\,)}, \quad i \in \text{supp}(\, \mathcal{S}^{[t]} \,)
\end{aligned}    
    \label{eqn:util-rough}
\end{equation}
Calculation of a segment's information utility depends on its attributes such as uncertainty, frontier score, and manipulability, all of which revolve around the Dynamic GPIS scene model.
}

\section{Dynamic Gaussian Process Implicit Surface}
\label{sec:dgp}
%Our mapping module utilises GP for arbitrary point inference, which becomes critical in dynamic scene point detection. Using the online GPIS fusion approach similar to \cite{bhoram-lee}, our module consists of two GP phases. We train GPIS from fused multi-scan measurements to infer scene surface, and use frame level GPR to detect changed GPIS training points. The difference to \cite{bhoram-lee} is dynamic scene handling and introduction of virtual wall.

The proposed dynamic mapping representation is based on the online GPIS fusion in~\cite{bhoram-lee}. In a similar way, our mapping module consists of two GP phases. A GPIS accumulated from multiple scans, and a frame-level 2.5D$^{-1}$ map as a detector for points removed from the scene. \textcolor{Blue}{It also exploits independent clusters to store GPIS training points for parallel processing.} The main difference with \cite{bhoram-lee} is the dynamic scene handling and introduction of a virtual wall as described in this section.

%\subsection{Continuous PIS as 3D Map Representation}
\subsection{Continuous distance function as 3D map representation}
\label{sec:gpis}
We now review how GPIS describes scene maps.
Let us define a point $\mathbf{x} \in \mathbb{R}^3$ and a function $f_{\text{IS}} : \mathbb{R}^3 \rightarrow{} \mathbb{R}$ such that
\begin{equation}
f_{\text{IS}} = 
\begin{cases}
+d \qquad \text{outside surface}\\
0  \qquad \text{on surface}\\
-d \qquad \text{inside surface}
\end{cases}
\label{eqn:gpis}
\end{equation}
where $d$ is the point-to-surface distance. The GPIS is defined by the posterior distribution of the value of $f$ at an arbitrary testing point $\mathbf{x}_{*}$ given by $f(\mathbf{x}_{*}) \sim \mathcal{N}(\bar{f}_{*},\mathbf{P}\left[f_{*}\right])$, where the predictive mean and variance are given by
\begin{equation}
\begin{aligned}
f_\ast     &= \mathbf{k}_\ast^\top( K + K_x )^{-1}\mathbf{y} \\
\mathbf{P}[f_\ast] &= k(\mathbf{x}_\ast, \mathbf{x}_\ast) - \mathbf{k}_\ast^\top(K + K_x)^{-1}\mathbf{k}_\ast
\end{aligned}
\label{eqn:gp-infer}
\end{equation}
$\mathbf{k}_\ast$, $K$ and $k(\mathbf{x}_\ast, \mathbf{x}_\ast)$ represent covariances between $\mathbf{x}_\ast$ and $n$ training points and $n\times n$ covariance matrix of the training points and the covariance function at $\mathbf{x}_\ast$, respectively~\cite{gp-book}. We use the Mat\'{e}rn $3$ class covariance function $(\nu=3/2)$,
\begin{equation}
k_m(d)\rvert_{\nu=3/2}  = (1 + \frac{\sqrt{3}d}{l})\exp(-\frac{\sqrt{3}d}{l}), \quad d = \|\mathbf{x}-\mathbf{x}'\|
\label{eqn:matern}
\end{equation}

\subsection{Instantaneous Scan \textcolor{Blue}{2.5D$^{-1}$} Map}
\label{sec:observation_GP}
%For each incoming depth image, a stand-alone elevation map is trained in the form of bearing angles $\bm{\theta}$ to inverse depth (IDP) GP.
For each incoming depth image, a stand-alone elevation map GP is created in the form of bearing angles $\bm{\theta}$ to inverse depth (IDP) regression. 
We call it a \textcolor{Blue}{2.5D$^{-1}$} map in this paper: 
\begin{equation}
    f_{\text{IDP}} : \textcolor{Blue}{\bm{\theta}} \rightarrow r^{-1}, \quad \textcolor{Blue}{\bm{\theta}=[\theta_u,\theta_v]^T}
    \label{eqn:gp-obs}
\end{equation}
where $r$ is measurement range. IDP is chosen for its i.i.d. Gaussian noise distribution  $\eta_{\text{IDP}} \sim \mathcal{N}( 0, \sigma^2_\text{IDP} )$. Note that every over-limit depth value should be replaced by a large user-defined number. %, to be explained in \ref{}{sec:gp-fuse}.\\ 
As in~\cite{bhoram-lee}, the Ornstein-Uhlenbeck (OU) covariance function is used to model IDP observations, 
\begin{equation*}
k_{\text{OU}}(d) = \frac{1}{2\alpha}\exp(-\alpha d), \quad d = \|\mathbf{x}-\mathbf{x}'\|
\end{equation*}
since the OU kernel is best suited for modelling random walk curves \cite{gp-book} without excessive smoothing.

Given the bearings $\textcolor{Blue}{\bm{\theta}}_s$,  one can infer its inverse depth $r_{\text{IDP}}(\textcolor{Blue}{\bm{\theta}}_s)$ and uncertainty $\sigma_{\text{IDP}}(\textcolor{Blue}{\bm{\theta}}_s)$  pair, and obtain coordinates of the corresponding point $\mathbf{x}_s$
\begin{equation}
    \begin{aligned}
    (&r_{\text{IDP}}, \sigma_{\text{IDP}}) = f_{\text{IDP}}(\textcolor{Blue}{\bm{\theta}}_s), \\
    \mathbf{x}_{s} &=  \frac{1}{r_{\text{IDP}}}{{\mathbf{v}}}, \quad {\mathbf{v}} = \frac{\mathbf{v}_h}{\|\mathbf{v}_h\|}, \quad
    \mathbf{v}_h = \begin{bmatrix}\textcolor{Blue}{\bm{\theta}}_s & 1\end{bmatrix}^\top\\
    \end{aligned}
    \label{eqn:r-IDP}
\end{equation}
The uncertainty $\sigma_{\text{IDP}}$ obtained defines a tolerance blanket of allowed range values from the sensor's viewing pose. Any test point in the field of view (FOV), yet falling outside the blanket, is considered an anomaly. This property will be exploited to detect removed objects in Section~\ref{sec:gp-fuse}.

\subsection{Data fusion in Dynamic GPs}
\label{sec:gp-fuse}
%The robot receives new view-point data as it moves around the environment. 
In this stage, the GPIS map is updated with new scan data. We delete samples from the existing GPIS training set that fall outside the tolerance layers defined by the \textcolor{Blue}{2.5D$^{-1}$} map. Then fuse or insert new scan points to the GPIS. This results in an immediately clean GPIS data structure that can be conditioned to infer the expanded environment map. The procedure is described in Algorithm \ref{Alg:dynamic-update}.  

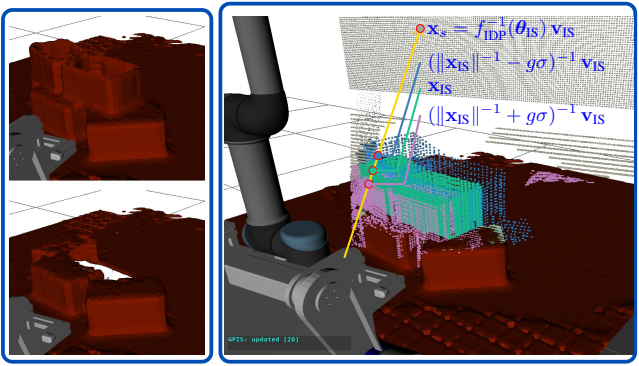
\begin{figure}[t]
	\centering
	\setlength\tabcolsep{1pt}
	\begin{tabular}{cc}	
	\tcbox[top=-1pt,left=-1pt,right=-1pt,bottom=-1pt,colback=white,colframe=blue!70!green]{
	\includegraphics[height=4.5cm,width=0.3\linewidth]{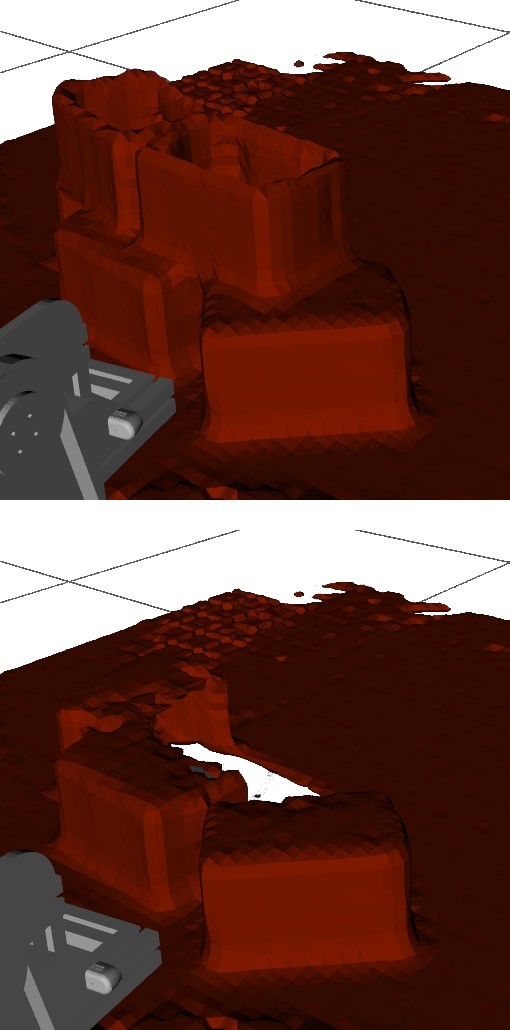}}
	&
%\tcbox[top=-1pt,left=-1pt,right=-1pt,bottom=-1pt,colback=white,colframe=gray]{\includegraphics[height=4.5cm,width=0.63\linewidth]{./figures/virtual_wall/virtual-wall-color_modified.jpg}}
\tcbox[top=-1pt,left=-2pt,right=-3pt,bottom=-1pt,colback=white,colframe=blue!70!green]{
\begin{tikzpicture}
\node[anchor=south west,inner sep=0] at (0,0) {\includegraphics[height=4.5cm, width=0.63\linewidth]{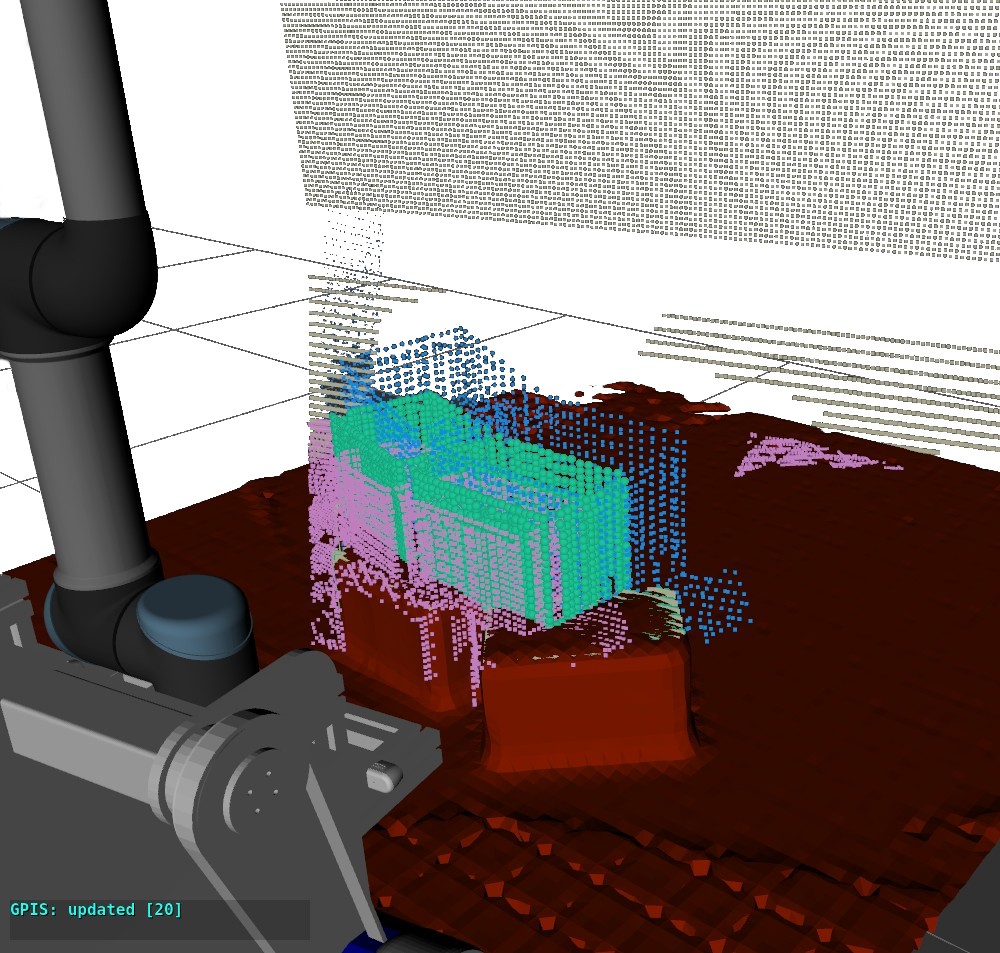}};
\draw[gold, thick, ->] (1.6,1.3) -- (2.61, 4.35) ;
\filldraw [color=red, fill=yellow_gray, thin] (2.61,4.35) circle (1.5pt) node[right, blue, scale=0.75] {$\mathbf{x}_{s} = f^{-1}_{\text{IDP}}({\bm{\theta}_{\text{IS}}})\,{\mathbf{v}}_{\text{IS}}$};
\filldraw [color=red, fill=teal, thin] (2.05,2.66) circle (1.5pt);
\filldraw [color=red, fill=pale_green, thin] (1.98, 2.46) circle (1.5pt) ;
\draw [color=red, fill=violet, thin] (1.925, 2.28) circle (1.5pt);
\draw [teal, thick](2.05,2.66) -- (2.24, 2.66) -> (2.61, 3.9) node[right, blue,scale=0.75] {$ (\|\mathbf{x}_{\text{IS}}\|^{-1} -  g\sigma_{} )^{-1}\,{\mathbf{v}}_{\text{IS}}$};
\draw [pale_green, thick](1.98,2.46) -- (2.30, 2.46) -> (2.61, 3.55) node[right, blue,scale=0.75] {$ \mathbf{x}_{\text{IS}}$};
\draw [violet, thick](1.925,2.28) -- (2.4, 2.28) -> (2.61, 3.2) node[right, blue,scale=0.75] {$ (\|\mathbf{x}_{\text{IS}}\|^{-1} +  g\sigma_{} )^{-1}\,{\mathbf{v}}_{\text{IS}}$};
\end{tikzpicture}
    }
	\\
	\small{\begin{tabular}{@{}l@{}}(\emph{a}) Map before and\\ after object remove\end{tabular}}
	&
	\textcolor{Blue}{\small{\begin{tabular}{@{}l@{}}(\emph{b}) Delete point $\mathbf{x}_{\text{IS}}$ if range significantly \\ different from its  \textcolor{Blue}{2.5D$^{-1}$} correspondence:\\ $\|\mathbf{x}_{\text{IS}}\|^{-1} - f_{\text{IDP}}(\frac{\mathbf{x}_{\text{IS}}}{\|\mathbf{x}_{\text{IS}}\|}) \gg \sigma_{\textrm{IDP}}$.	\end{tabular}}}
\end{tabular}			
	\caption{\small{\textcolor{Blue}{Map update in Dynamic GP. Using ray-casting a reverse tolerance layer is formed around GPIS points: $+\sigma_{\textrm{IDP}}$ (\textcolor{teal}{blue}) and $-\sigma_{\textrm{IDP}}$ (\textcolor{violet}{violet}). Removed object points (\textcolor{pale_green}{green}) are detected and deleted from GPIS, forming a new map. The \textcolor{Blue}{2.5D$^{-1}$} map with the virtual wall at ``infinity'' is shown in}  \textcolor{yellow_gray}{dark yellow}. \textcolor{Blue}{An example FOV ray is shown in} \textcolor{yellow}{yellow}}}
	\label{fig:dynamic-delete}
{\vspace{-10pt}}		
\end{figure}

We first identify all GPIS samples that fall into the sensor's FOV at the current pose. For each point in GPIS we transform its coordinates to the local camera's frame $\mathbf{x}_{\text{IS}} = (\mathbf{T}^{[t]})^{-1} \mathbf{x}_{\text{IS}}^{(\textrm{w})}$. 
%With a given camera pose $\mathbf{T}^{[t]}\in \mathrm{SE}(3)$, we obtain local coordinates for each point in GPIS $\mathbf{x}_{\text{IS}} = (\mathbf{T}^{[t]})^{-1} \mathbf{x}_{\text{IS}}^{(\textrm{w})}$. 
The bearing angles $\textcolor{Blue}{\bm{\theta}}_{\text{IS}}$ are obtained after homogeneous normalisation. For points inside the FOV, we infer its IDP and uncertainty pair $(r_{\text{IDP}}, \sigma_{\text{IDP}} )$ as described in Section~\ref{sec:observation_GP}.
%Specifically, for any point $\mathbf{x}_{\text{IS}}^{(\textrm{w})}$ in the GPIS training set, we obtain its local coordinates $\mathbf{x}_{\text{IS}} = (\mathbf{T}^{[t]})^{-1} \mathbf{x}_{\text{IS}}^{(\textrm{w})}$, where $\mathbf{T}^{[t]}=(R,\mathbf{t})\in \mathrm{SE}(3)$ is the sensor's pose and is assumed given. We use the bearings $ [\bm{\theta}_{\text{IS}} \; 1]^T=\pi({\mathbf{x}_{\text{IS}}})$ to select FOV samples. From the valid angles we infer its IDP and uncertainty pair $(r_{\text{IDP}}, \sigma_{\text{IDP}} )$ as described in \ref{sec:observation_GP}. 
Then we examine the difference $\triangle_{\text{IS}}$ between the stored inverse depth $r^{-1}_{\text{IS}} = \|\mathbf{x}_{\text{IS}}\|^{-1}$ and the inferred one $r_{\text{IDP}}$:
\begin{equation}
    \triangle_{\text{IS}} = {r}^{-1}_{\text{IS}} - r_{\text{IDP}}(\textcolor{Blue}{\bm{\theta}}_{\text{IS}}) \, ,
    \label{eqn:delta_inv}
\end{equation} 
to apply three possible treatments: delete, fuse or ignore. \\
\noindent \textbf{Delete:} For regions containing removed objects, the new scan reveals what is lying behind the old object along its light ray. The range measured would be either larger or outside the sensor range. We delete this point using the anomaly detection criterion, 
%\begin{equation}
 $   \triangle_{\text{IS}} \ge g\, \sigma_{\text{IDP}} $,
%\end{equation}
where $g$ is a constant scale factor. 
However, this simple check fails if the background is empty since it destroys the 2.5D$^{-1}$ map model. Here, using a method similar to \cite{simon-acra}, we artificially replace every over-limit measurement in the scanned depth with a very large number, effectively creating a virtual wall at ``infinity'', see Fig. \ref{fig:dynamic-delete} for illustration. The virtual wall is processed together with the rest of the scan data to form the \textcolor{Blue}{2.5D$^{-1}$} map, allowing valid inferences for every bearing angle. 

\noindent \textbf{Fuse:} For $\triangle_{\text{IS}}$ within a small range, the old GPIS point $\mathbf{x}_{\text{IS}}$ will be fused with its corresponding point on the \textcolor{Blue}{2.5D$^{-1}$} surface as described in \cite{bhoram-lee}.\\ %, the reader is encourage to refer to  \cite{bhoram-lee} for details.\\
\noindent \textbf{Ignore:} Points outside the above categories are occluded scene points and should be left unchanged. 
\begin{algorithm}[ht]
\SetAlgoLined
\SetNlSty{textbf}{}{:}
\nl $f_{\mathrm{IDP}} \leftarrow \textcolor{Blue}{\text{Regress2.5D}^{-1}}( \text{Scan}^{[t]} ) $\;
\nl \For{ $\mathbf{x}_{\mathrm{IS}}^{(\mathrm{w})} \in \mathrm{GPIS}^{[t]}$}{
     $\mathbf{x}_{\text{IS}} = (\mathbf{T}^{[t]})^{-1} \, \mathbf{x}_{\text{IS}}^{(\mathrm{w})}$\;
     $(r_{\text{IDP}}, \sigma_{\text{IDP}}) \leftarrow f_{\text{IDP}}(\,\textcolor{Blue}{\bm{\theta}}_{\text{IS}}\,) $ \tcp*{ \Scale[0.70]{\texttt{Eq. (\ref{eqn:gp-obs})}}}
	 $\triangle_{\text{IS}} \leftarrow \text{Compare}( f_{\text{IDP}}, \|\mathbf{x}_{\text{IS}}\|^{-1} )$ \tcp*{\Scale[0.70]{\texttt{Eq. (\ref{eqn:delta_inv})}}}
    \eIf{$\triangle_{\mathrm{IS}} \ge g \sigma_{\mathrm{IDP}}$}
    	{ %the old point must have been taken away from the scene for the see-through effect, so delete 
    	\tcc{\Scale[0.90]{\texttt{point removed from scene}}}
    	$ \mathbf{x}_{\text{IS}}^{(\mathrm{w})} \leftarrow \varnothing$ \;
    	}
    	{
%        	\eIf{ $ -\sigma_{occ} \le \triangle_{\text{IS}} \le \sigma_{occ} $}
        	\If{ $ -g\sigma_{\mathrm{IDP}} \le \triangle_{\mathrm{IS}} \le g\sigma_{\mathrm{IDP}} $}  
        	{
        	\tcc{\Scale[0.90]{\texttt{old and new fusable}}}
        	$\mathbf{x}_s \leftarrow \text{Invert}(r_{\, \text{IDP}}, \textcolor{Blue}{\bm{\theta_}{\mathrm{IS}}} \,)$ \tcp*{ \Scale[0.70]{ \texttt{Eq.  (\ref{eqn:r-IDP})}} }
        	 $\mathbf{x}_{\text{IS}}^{(\mathrm{w})} \leftarrow {\mathbf{T}^{[t]}} \, \mathrm{Fuse}( \mathbf{x}_{\mathrm{IS}}, \mathbf{x}_s)$ \tcp*{ \Scale[0.7]{ \textcolor{Blue}{\texttt{Ref. \cite{bhoram-lee}}}}} 
        	}
        }
	}
	%\nl Insert the new scan points into the data structure  train
	\nl $\text{GPIS}^{[t+1]} \leftarrow \mathrm{GPIS}^{[t]} \bigcup \mathrm{Scan}^{[t]}$\; 
\caption{Dynamic GPIS Update}
\label{Alg:dynamic-update}
\end{algorithm}

\textcolor{Blue}{With the old GPIS training points cleaned up, the new scan samples are now added for map expansion. The modified local GPIS clusters from both ends are processed further (covariance inversion) for fast future inference. For a full treatment of online GPIS formulation please refer to~\cite{bhoram-lee}.}

\section{NBV Selection}
\label{sec:nbv}
To select an optimal next pose, we first identify a set of candidate poses, then compute its utility function, incorporating various GPIS based metrics including gradient frontier and manipulability. 
%\todo[inline]{Re-organize and merge A. \& B.}
\subsection{Candidate Poses}
\label{sec:candidate}
\begin{figure}[h]
	\centering
	\setlength\tabcolsep{1pt}	
	\begin{tabular}{ll}	
	\tcbox[arc=1pt,top=-1pt,left=-1pt,right=-1pt,bottom=-1pt,colback=white,colframe=blue!70!green]{\includegraphics[clip, trim=0cm 1.5cm 0cm 0cm,width=0.46\linewidth]{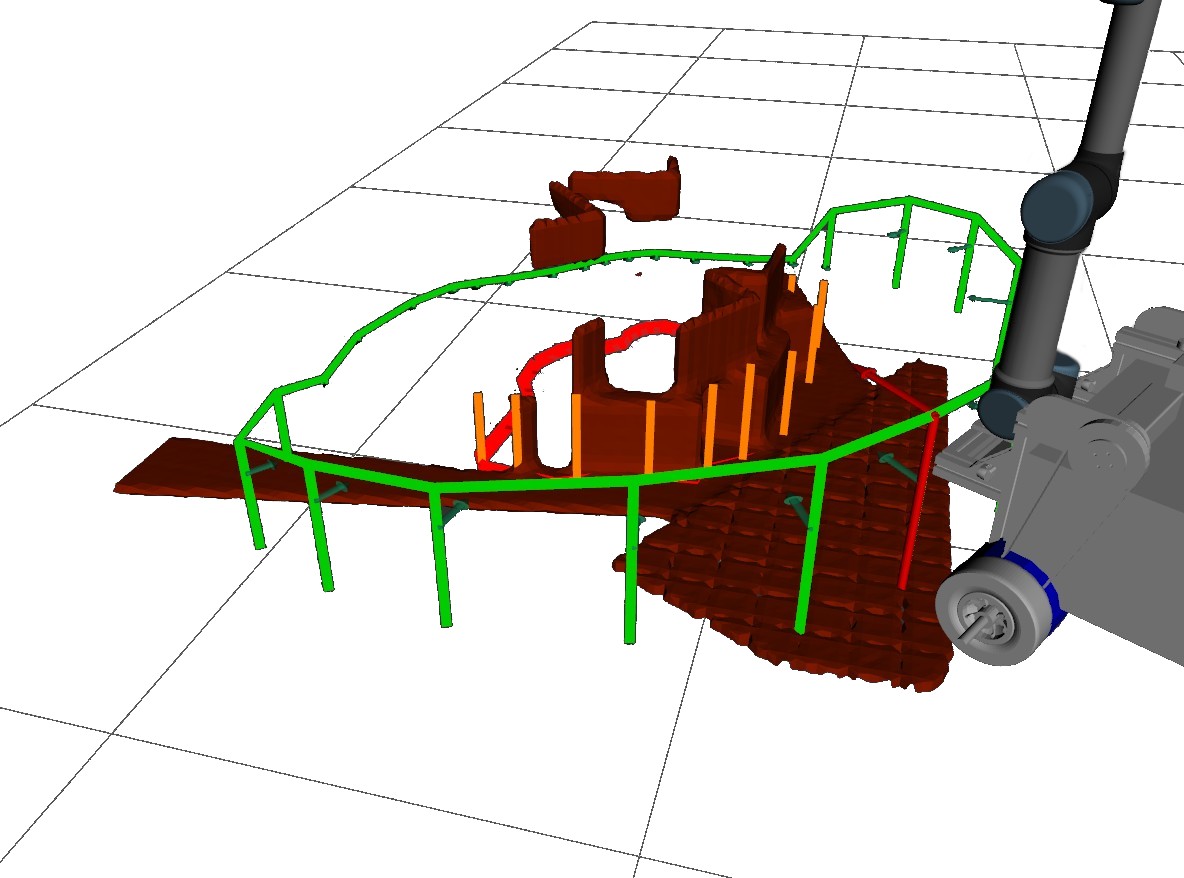}} &
	\tcbox[arc=1pt,top=-1pt,left=-1pt,right=-1pt,bottom=-1pt,colback=white,colframe=blue!70!green]{\includegraphics[clip, trim=0cm 1.5cm 0cm 0cm,width=0.46\linewidth]{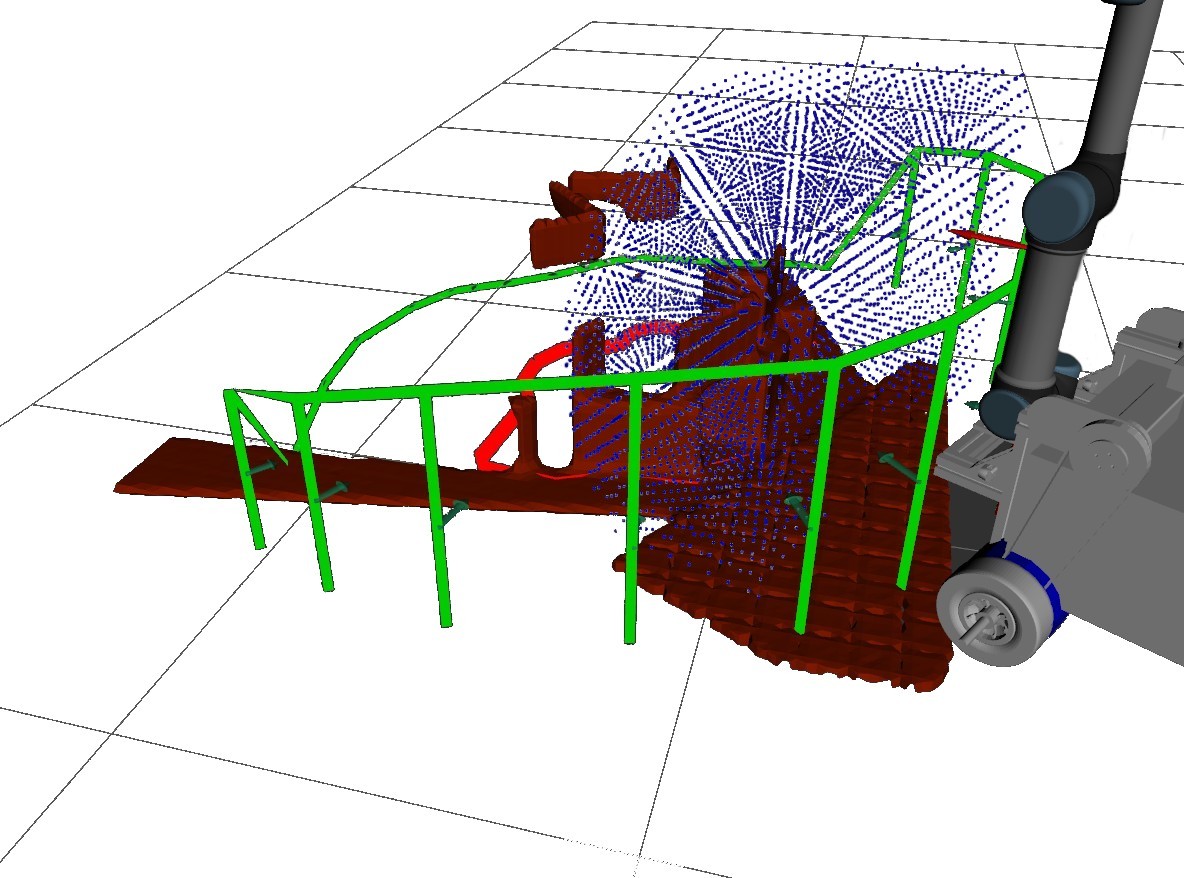}}  \\[-7pt]
\small{ (\emph{a}) Complete utility} 	
& 	\small{(\emph{b})  Manipulability factor } \\
	\tcbox[arc=1pt,top=-1pt,left=-1pt,right=-1pt,bottom=-1pt,colback=white,colframe=blue!70!green]{\includegraphics[clip, trim=0cm 1.5cm 0cm 0cm,width=0.46\linewidth]{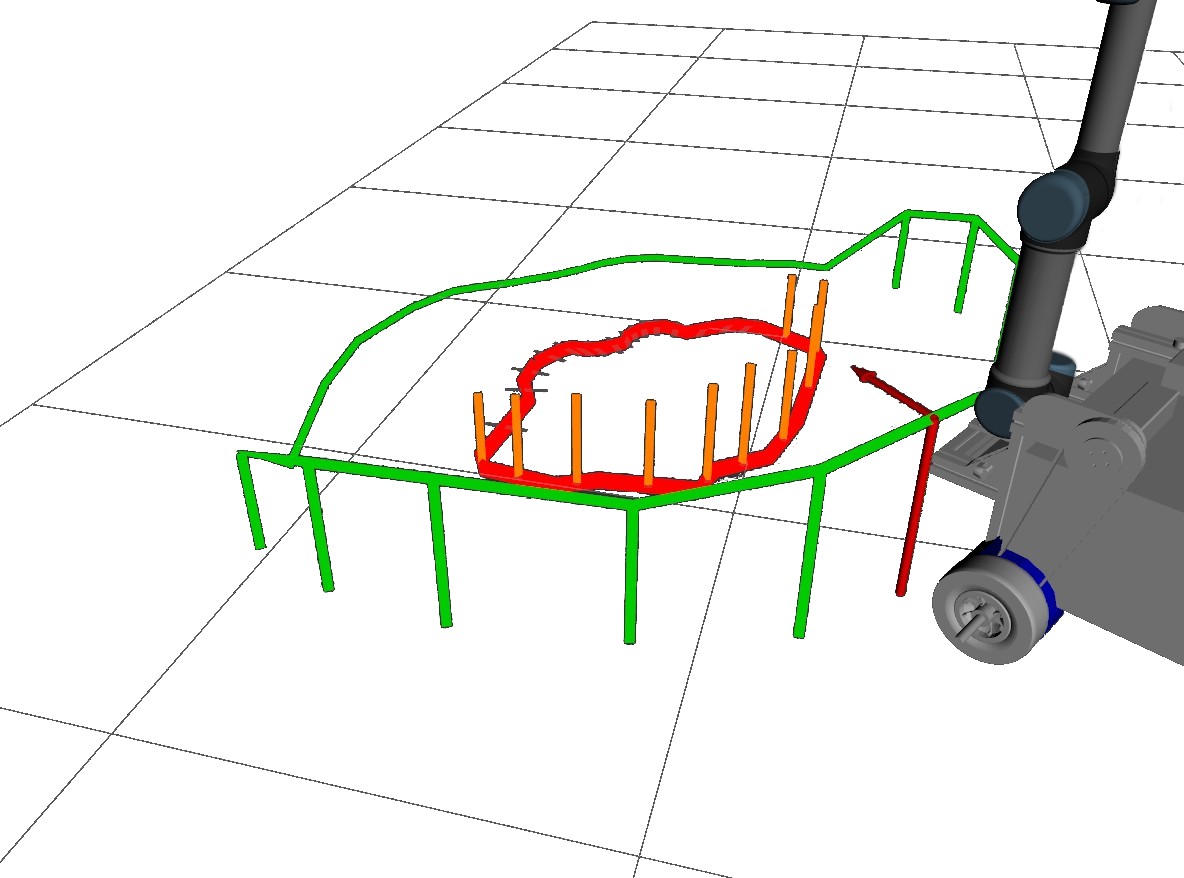}} &
	\tcbox[arc=1pt,top=-1pt,left=-1pt,right=-1pt,bottom=-1pt,colback=white,colframe=blue!70!green]{\includegraphics[clip, trim=0cm 1.5cm 0cm 0cm,width=0.46\linewidth]{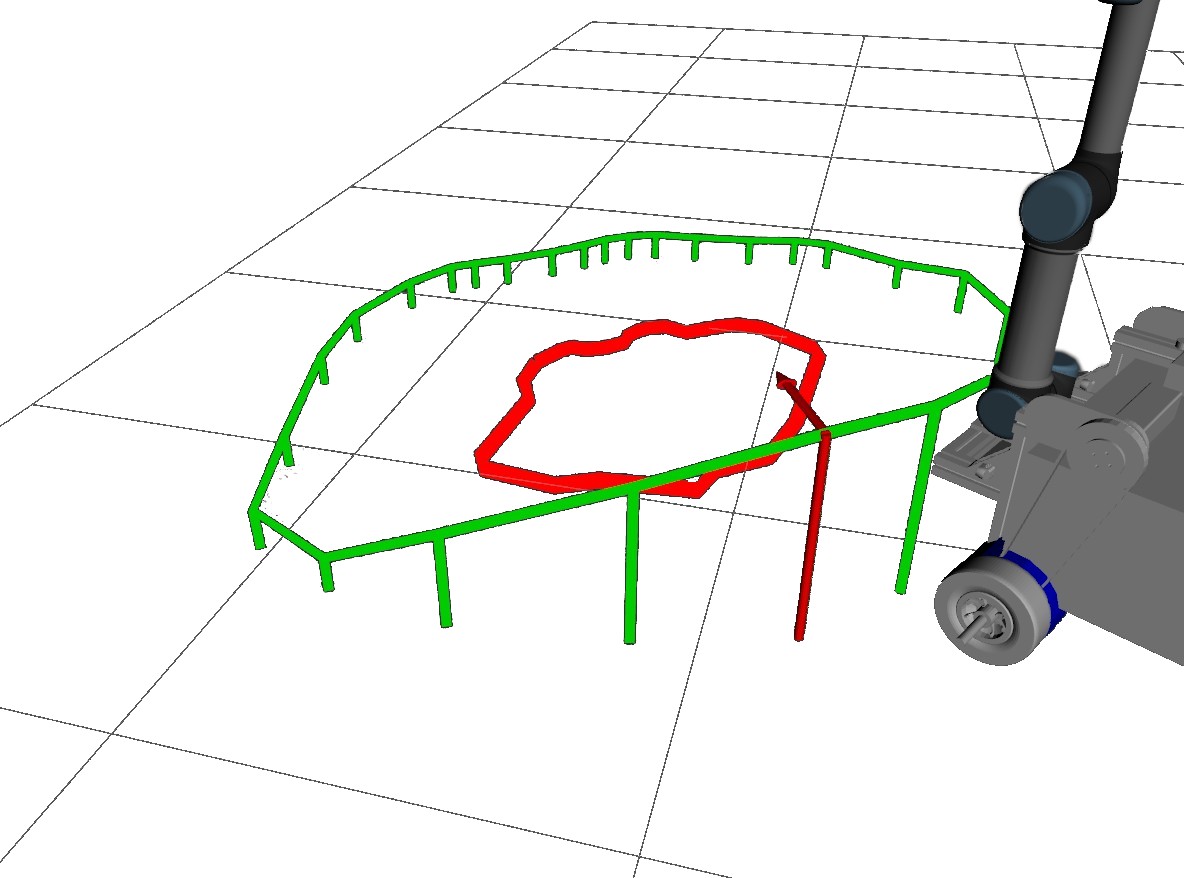}}  \\[-7pt]
	\small{\emph{c})    Interact order factor} 
		& 	\small{(\emph{d}) Travel distance factor } \\
	\tcbox[arc=1pt,top=-1pt,left=-1pt,right=-1pt,bottom=-1pt,colback=white,colframe=blue!70!green]{\includegraphics[clip, trim=0cm 1.5cm 0cm 0cm,width=0.46\linewidth]{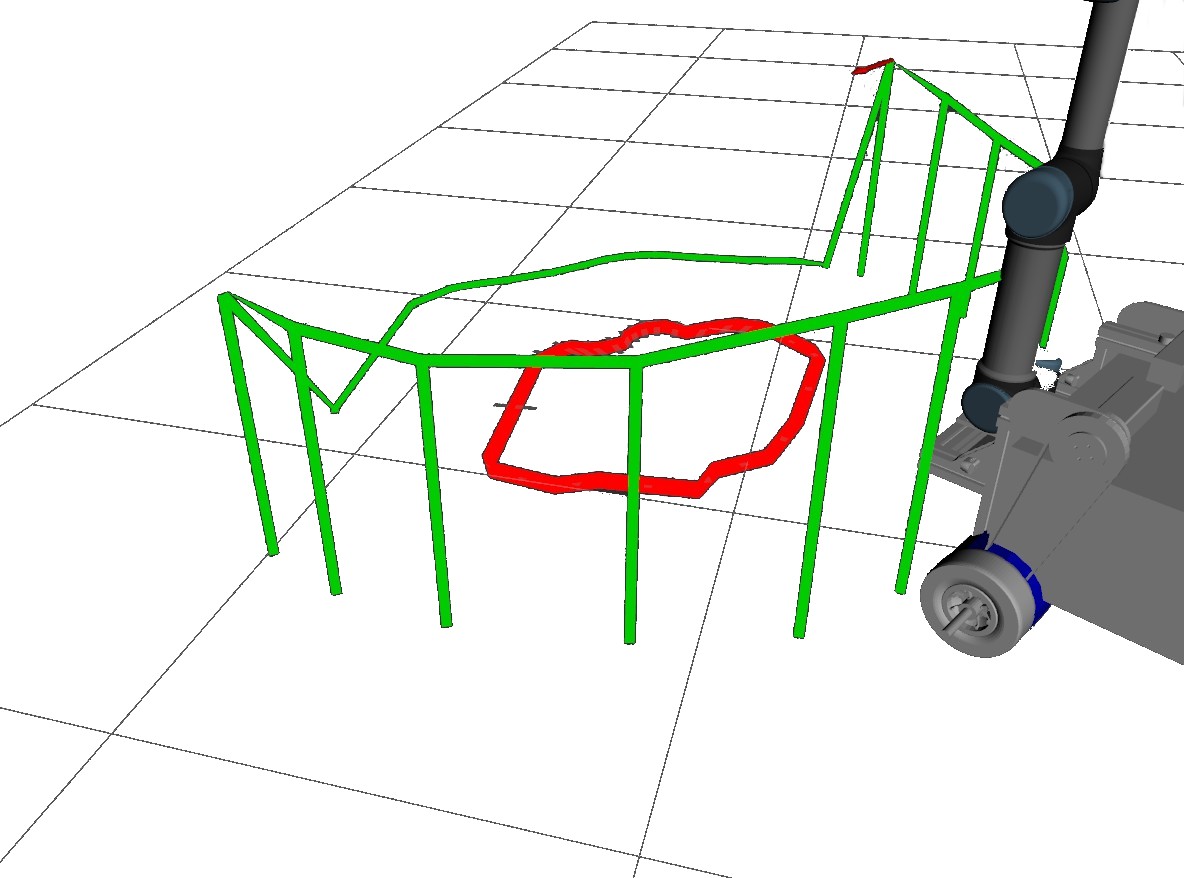}} &
	\tcbox[arc=1pt,top=-1pt,left=-1pt,right=-1pt,bottom=-1pt,colback=white,colframe=blue!70!green]{\includegraphics[clip, trim=0cm 1.5cm 0cm 0cm,width=0.46\linewidth]{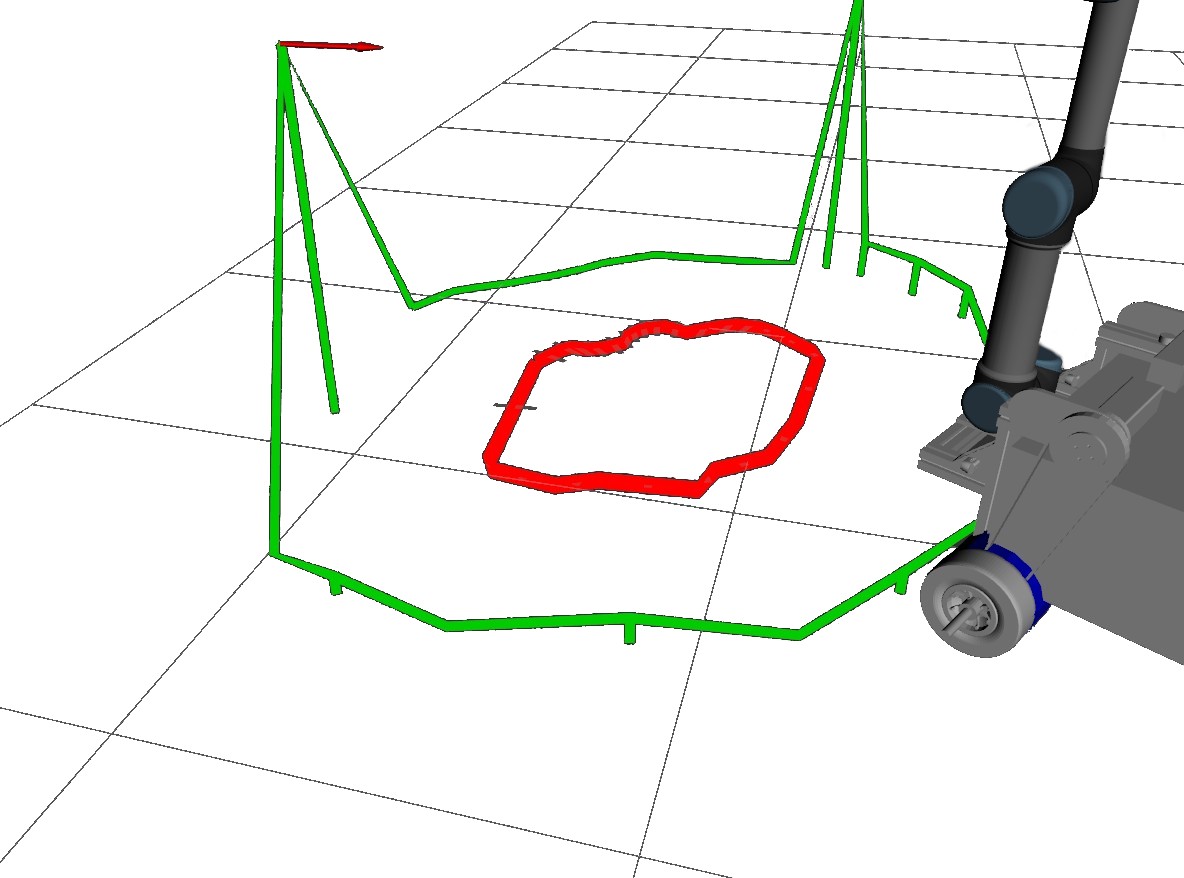}}  \\ [-7pt]
\small{(\emph{e}) Uncertainty factor } 	
& 	\small{(\emph{f}) Frontier factor}
\end{tabular}			
	\caption{\small{An illustration of full formulation and single factor utilities. First, identify pile segments (red contour) and compute the utilities (green bars). Max utility gives NBV (red bar+arrow). (\emph{b}) shows samples (blue dots) from the manipulability annulus. (\emph{c}) shows height (orange bar). Grey bars on the ground are uncertainty for imaginary segments}.}
\label{fig:utility}
{\vspace{-10pt}}	
\end{figure}

Since the robot needs to explore objects in a pile, the next position to place the robot should be around the circumference of the partially explored pile, as shown in Fig. \ref{fig:utility}. We first query the GPIS to obtain a probabilistic implicit surface representing the scene, then use the marching cubes algorithm \cite{marching-cube} to form a dense map. The map can be assumed to have a roughly conical shape which is generically the case when stacking objects to maintain stability. We project the surface points onto the ground to form a 2D occupancy map. Using image processing methods, we obtain the contour of the occupancy, (see Fig. \ref{fig:utility}(\emph{a})), which is composed of a set of piece-wise linear segments $\{\mathbf{s}_i\}$ of the map base. These segments are the candidates for positioning the robot in the next step. Half of $\{\mathbf{s}_i\}$ are from the unseen surface area without full exploration, hence have high uncertainty, referred to as ``imaginary". The remaining half are genuine base outline segments and have low uncertainty. Using this uncertainty we classify each segment $\textbf{s}_i$ as real or imaginary. 

\subsection{Utility Formulation}
\label{sec:utility}
Many factors (or attributes) from $\mathbf{s}_i$ can affect the utility. We use $\{\cdot\}[\,\mathbf{s}_i^{[t]}\,]$ to denote attribute $\{\cdot\}$ of segment $i$. For the sake of simplicity, from now on we omit the time superscript in segment attributes. These attributes are:
\begin{itemize}
\item \textbf{Uncertainty $\sigma^2[\mathbf{s}_i]$}:
encourages information collection for noisy regions. Only the real segments are considered here. The imaginary ones are addressed below.

\item \textbf{Frontier $f_{\nabla\sigma^2}[\mathbf{s}_i]$}: for unexplored regions, frontier gives preference to the boundary segments over other imaginary segments. 
%See \ref{sec:sigma_gradient} for details, the contour segment becomes the transversal line segment described therein.

\item \textbf{Arm manipulability $m[\mathbf{s}_i]$}: This factor gives a quantitative measure of volume in GPIS that is reachable by the manipulator. 
%See  \ref{sec:interact}.

\item \textbf{Interact order $h[\mathbf{s}_i]$}:
gives a preference for picking order and can be in the form of segment height, readily available from the GPIS. This order factor is useful since picking in an order from high to low positions causes minimum disruption to the object pile. Further, this encourages the pile to maintain a convex-shaped outline, convenient for base navigation. 

\item \textbf{Travel distance $d[\mathbf{s}_i]$}:
preference is given to visit close-by scanned segments. It is computed as the Dijkstra minimum distance from current location to candidate.

\item \textbf{Avoid repeated failure $p(\, \mathbf{s}_i, t \,)$}: avoids locations where previous pick/detect attempts failed within a time duration, and is a function of time. 
% See \ref{sec:candidate}. 
\end{itemize}

Considering these factors, we present the following utility formulation to select the optimal candidate segment ${i_*}$:
\begin{equation}
	\begin{aligned}
		i_*^{[t]} &=  \underset{i}{\text{argmax}} \, U^{[t]}(\,\mathbf{s}_i\,), \quad i \in \text{supp}(\, \mathcal{S}^{[t]} \,), \\
		U^{[t]} (\, \mathbf{s}_i \,) &= (1 - p(\,t\,)[\mathbf{s}_i] )\; \mathcal{I}(\,\mathbf{s}_i\,), \\
	\mathcal{I}(\,\mathbf{s}_i\,) &= \beta_1 \, \mathcal{L}_1(\,m[\mathbf{s}_i]\,) 
		 + \beta_2 \, \mathcal{L}_2(\,h[\mathbf{s}_i]\,) \\
		 &+ \beta_3 \, \mathcal{L}_3(\,-d[\mathbf{s}_i]\,) 		 +  \beta_4 \, \mathcal{L}_4(\,\sigma^2[\mathbf{s}_i]\,) \\
		 &+ \beta_5 \, \mathcal{L}_5(\,f_{\nabla\sigma^2}[\mathbf{s}_i]\, ),\\
		1 &= \sum_{\textcolor{Blue}{k}} \beta_{\textcolor{Blue}{k}}, \quad \textcolor{Blue}{\beta_k \ge 0}, \\%\quad k \in [1..5]\\
		%\beta_1 + \beta_2 + \beta_3 + \beta_4 + \beta_5, \\
		p(\,t\,)[\mathbf{s}_i] &=   \textcolor{Blue}{\gamma}^{\, t - t_f[\mathbf{s}_i] \,}, \quad \textcolor{Blue}{\gamma < 1} \\ 
		\mathcal{S}^{[t+1]}  & \leftarrow \text{action}(\  i_*^{[t]} \  )
		\label{eqn:util}
	\end{aligned}
\end{equation}
where 
\begin{equation}
	\begin{aligned}
\mathbf{s}_i    &:= ( \,
    m, \,
    \sigma^2, \, 
    f_{\nabla\sigma^2}, \,
    h, \,
    d, \,
    p(t) \,
    )[\mathbf{s}_i], \\
	   \mathcal{I}(\cdot) &: \text{information gain} \\
		\forall \, k &= 1,2,3,4,5 \\
		\beta_k &: \text{ user defined weightings}  \\
		\mathcal{L}_k(\,x\,) &: \text{ logistic function, maps to probability}\\
		&  = \frac{l_k}{1 + \text{exp} (\, -a_k x + b_k \,) }, \\
		(l_k, a_k, b_k) &: \text{empirically obtained parameters} \\
	\end{aligned}
\end{equation}

%Due to cluttering or plan failure, a robot at a seemingly optimal position may fail to detect or pick up objects. Yet once the neighbouring regions are cleared, the location may be useful again. As seen in Eq. (\ref{eqn:util}), a \textit{discounted} future penalty $p(\mathbf{s}_i, t)$ on the information gain $\mathcal{I}(\mathbf{s}_i)$ ensures the cell is penalised and excluded from NBV selection for a small timeframe immediately after failure.
%For this to work, a quad-tree data structure stores the last failure time $t_f$ at each discrete location. Initially, all locations have $t_f$ set to negative infinity, equivalent to no penalty. Once a failure has occurred, the $t_f$ at position $\mathbf{s}_i$ is updated to the current time.  
\noindent Note that $p(\mathbf{s}_i, t)$ takes the form of \textit{discounted} future penalty. It relies on the duration between the last failure time $t_f$ and the current time. Initially we set $t_f$ to $-\infty$ for zero penalty and update it once a failure occurs. This activates the penalty and deactivates the segment for a small time frame. 
\\
During runtime, all imaginary segments have their manipulability, height, and uncertainty set to zero, resulting in zero utility. This prevents the robot from landing on unknown regions. The frontier segments with valid $m$, $h$ and $\sigma^2$ scores, compete with other real segments for NBV selection. Once the frontier is explored, its surrounding imaginary regions become ``real'' with the frontier shifted. The new segments are added to the next round of NBV selection.  
 
 %choosing the optimal segment $i_*$ at each time instance $t$: 

 \subsection{Gaussian Process Frontier}
\label{sec:sigma_gradient}
We now propose our GP based frontier metric which gives higher precedence for boundary segments over unexplored regions. The uncertainty gradient for an arbitrary point $\mathbf{x}$ can be defined as
\begin{equation*}
    \| \nabla \sigma^2(\mathbf{x} ) \|^2. 
{\vspace{-50pt}}    
\end{equation*}
The robot base motion is confined to the 2D ground plane, during manipulation the base is aligned width-wise with the pile segment. Due to the segment's finite length, it is necessary to integrate the gradients along the segment direction to obtain the overall uncertainty variation.
\\
%it is incorrect to approximate the frontier as a point gradient in 3D space. Rather, we use a line segment to define the space occupied by the robot, then retrieve the ground plane GPIS frontier. 
Let $\mathbf{l}_i = [l_x, l_y, 0]^\top$ denote the direction of segment $\mathbf{s}_i$, transversal to the robot heading.
Let $\mathcal{P}_i$ denote the set of 3D points in $\mathbf{s}_i$. Denote $\delta \alpha$ as the infinitesimal segment length along $\mathbf{l}_i$. Then for a point $\mathbf{x}_j^i \in \mathcal{P}_i$, a small perturbation results in a neighbour point $\mathbf{x}_j^i + \delta \alpha \mathbf{l}_i$. We define the frontier metric as the Sum of Directional Squared Difference (SDSD) in uncertainty for all points in $\mathcal{P}_i$ along $\mathbf{l}_i$:
%$\[\Scale[0.5]{y = \sin^2 x}\]$
\begin{equation}
    \Scale[0.90]{
\begin{aligned}
f_{\nabla \sigma^2}  &= 
    \sum_{\mathbf{x}_j^i \in \mathcal{P}_i}  \biggm\Vert \, \frac{\partial  (\, \sigma^2(\mathbf{x}_j^i + \delta \alpha \mathbf{l}_i )  - \sigma^2(\mathbf{x}_j^i ) \, ) }{\partial \delta\alpha} \biggm\Vert^2 
      \\
    & = \sum_{\mathbf{x}_j^i \in \mathcal{P}_i}  
    \biggm\Vert \, \frac{\partial  (\, \sigma^2(\mathbf{x}_j^i) + \delta \alpha \nabla \sigma^2(\mathbf{x}_j^i)^\top \, \mathbf{l}_i   - \sigma^2(\mathbf{x}_j^i ) \, ) }{\partial \delta\alpha} \biggm\Vert^2  \\
    & = \sum_{\mathbf{x}_j^i \in \mathcal{P}_i}   \biggm \| \nabla \sigma^2(\mathbf{x}_j^i ) ^\top \, \mathbf{l}_i \biggm \|^2  
    %&=  \sum_{j \in \mathbb{S}_i} \mathbf{l}_i ^T (\nabla \sigma^2(\mathbf{x}_j^i )  \nabla \sigma^2(\mathbf{x}_j^i )^\top )    \mathbf{l}_i \\
\end{aligned}
}
\label{eqn:frontier}
\end{equation}
For gradient at point $\mathbf{x}_j^i$, we use GPIS definition (\ref{eqn:gp-infer}) and Mat\'{e}rn kernel (\ref{eqn:matern}) to compute as follows:
\begin{equation}
\begin{aligned}
    %f_{\nabla \sigma^2} \, =&  \, \sum_{j \in \mathbb{S}_i}
    %2 \, \biggm| \,     \mathbf{k}_*^T(K + K_x)^{-1} \nabla\mathbf{k}_*(\mathbf{x}_j^i) \,  \mathbf{l}_i \,
    %\biggm|   \\
    \nabla \sigma^2(\mathbf{x}_j^i )^\top \, =& \; 2 \,  \mathbf{k}_*^\top \, (K + K_x)^{-1} \,  \nabla\mathbf{k}_*(\mathbf{x}_j^i)^\top \\
    \nabla\mathbf{k}_*(\mathbf{x}_j^i)^\top \,  =&  \; \begin{bmatrix}  \, ..., & \nabla{k}_m (\mathbf{x}', \mathbf{x}_j^i)^\top, \, & ... \;
    \end{bmatrix}^\top\\
    %\text{let } r =& \, \| \mathbf{x}_j^i - \mathbf{x}'\| \\
    \nabla{k}_m (\mathbf{x}', \mathbf{x}_j^i )^\top \,  =& \, \frac{\partial {k}_m(d)}{\partial d}
    \frac{\partial d} {\partial \mathbf{x}^i_j} \\
\end{aligned}
\end{equation}
\begin{equation*}
\text{where:}\quad d = \,\biggm\Vert \, \mathbf{x}_j^i - \mathbf{x}' \, \biggm\Vert,  \quad
\frac{\partial{k}_m(d)    }{\partial d} =  \,
-\frac{3d}{l^2}\exp(-\frac{\sqrt{3}d}{l}).
%\,    \frac{\sqrt{3}}{l}\exp(-\frac{\sqrt{3}r}{l}) \\
%    & - \frac{\sqrt{3}}{l}(1+\frac{\sqrt{3}r}{l})\exp(-\frac{\sqrt{3}r}{l}) 
\end{equation*}

\textcolor{Blue}{This GP-based definition is continuous and mathematically rigorous. It contrasts to the discretised frontier metric in~\cite{entropy-gradient}. Its simple formulation is more tractable and deployable than~\cite{maani-gp}, which is only for 2D maps and relies on an auto-jacobian method from optimisation libraries, due to the hybrid GPOM world model used.}
\\
\subsection{Manipulability}
\label{sec:interact}
\begin{figure}[h]
    \centering
    \setlength\tabcolsep{1pt} 
    \includegraphics[height=6.0cm]{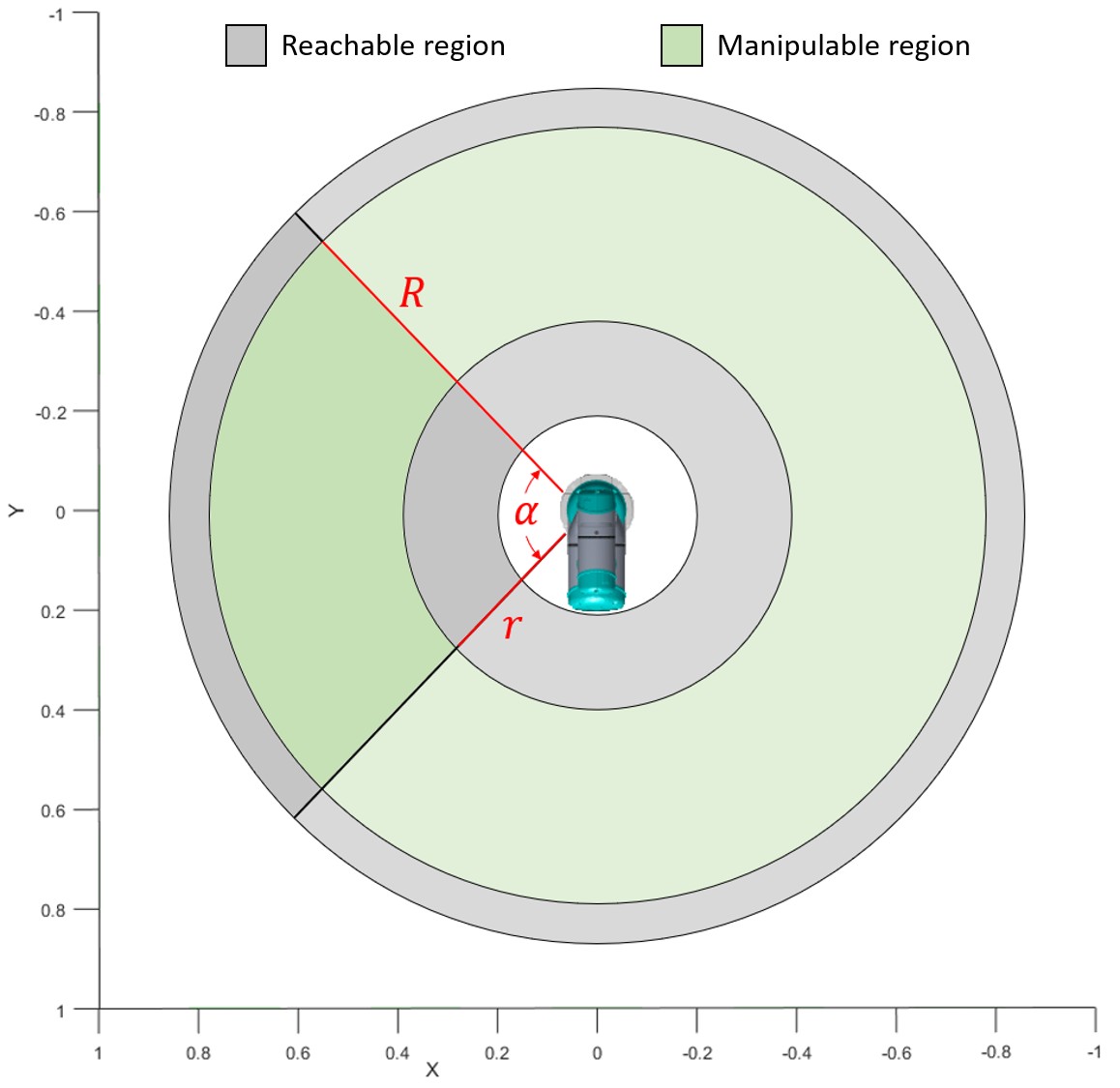}
    \caption{\small{A top-view of the annulus sector used in the pose manipulability filter.}}
    \label{figure:manip_filter}
\end{figure}
This section presents a metric to determine whether the manipulator is capable of reaching and picking the objects in the environment. 
%This is achieved by subjecting all bricks returned from the perception module to pose manipulability filtering.
This metric filters out object poses that minimise the manipulability index~\cite{Yoshikawa85} of the arm during picking. The filter uses the geometric volume of an annulus (donut-like) sector as a heuristic to determine which poses are viable. The frame of this region is fixed to the manipulator base frame and all poses outside this region are rejected by the filter. For this process, we use the determinant of the Jacobian matrix $\mathbf{J}$ to calculate the robot’s measure of manipulability $m$ as follows,
\begin{equation}
    m = \sqrt{\det (\mathbf{J} \mathbf{J}^T (\theta))} \in \mathbb{R}\,
\end{equation}
where a robot's Jacobian $\mathbf{J}$ relates an end-effector's Cartesian velocity with the joint velocities $\dot{\mathbf{q}}$. 
The annulus sector used in the pose manipulability filter is derived from the manipulator’s kinematics by sampling valid end-effector discrete poses. %To estimate the annulus sector a kinematic model of the UR5e was created in MATLAB using Peter Corke’s Robotics Toolbox \cite{corke2002robotics}.
%
%%The configuration space $\mathbb{C}_{Full}$ is defined as all the possible configurations of the manipulator's joints $\mathbf{q}$. %The configuration of the manipulator at any given instant is defined by the joint positions $\mathbf{q}$ where:
%\begin{equation}
%    \begin{aligned}
%        \mathbf{q} = \{q_j, \, %j \in \mathrm{ [1 \ldots N]} \} %\in \mathbb{C}_{Full} \\
%        \mathbb{C}_{Full} %\subset \mathbb{R}^N
%    \end{aligned}
%\end{equation}
%
%%Performing an analysis of the entire configuration space would be too computationally expensive. Thus, a discrete sampling $\mathbb{C}_{Sample} \subset \mathbb{C}_{Full}$ is obtained for all possible manipulator configurations by incrementing each joint by a fixed distance $\Delta \theta$ between the joint limit intervals $[\theta_{min}, \theta_{max}]$.
%
%%For every configuration $\forall \mathbf{q} \in \mathbb{C}_{Sample}$ we calculate the end-effector pose in task space $\mathcal{T}$ using forward kinematics $\xi_E = K(\mathbf{q})$, as well as the manipulability index $m$ where:
For each configuration $\mathbf{q}$ in the configuration space, we calculate the end-effector pose $\xi_E$ in task space $\mathcal{T}$ using forward kinematics %$\xi_E = K(\mathbf{q})$, 
and the manipulability index $m$ where:
\begin{equation}
    \begin{aligned}
        \xi_E &= (R, \mathbf{p}) \in \mathcal{T}, \quad \mathcal{T} \subset \mathrm{SE(3)}, \\
        %\mathbf{p} &= \begin{bmatrix} x & y & z \end{bmatrix}^T, \quad \mathbf{p} \in \mathbb{R}^3
    \end{aligned}
\end{equation}
with the position $\mathbf{p} \in \mathbb{R}^3$ and the rotation $R \in \mathrm{SO(3)}$.

%%This process gives us an n-by-4 matrix containing the end-effector position and manipulability index for each sampled configuration,
%
%%\begin{equation}
%%    \begin{aligned}
%%        A &= [ \mathbf{a}_1^T, \ldots, \mathbf{a}_i^T, \ldots, \mathbf{a}_n^T ]^T \in \mathbb{R}^{n\times4},\\
%%       \mathbf{a}_i &= [\mathbf{p}^T, m] \in \mathbb{R}^{1\times 4}
%%    \end{aligned}
%%\end{equation}
%
%%We discard any rows with a manipulability index $m$ below a threshold $m_{thres}$,
%
%%\begin{equation}
%%    \begin{aligned}
%%        A^{-} &= A \ \setminus A_I, \\
%%        A_I &= \{ \mathbf{a}_i \vert i \in 1 \ldots n \wedge \mathbf{a}_{i, 4} \le m_{thres} \}
%%    \end{aligned}
%%\end{equation}

We select only the configurations with manipulability index $m \geq m_{thres}$. The remaining points are used to determine the minimum and maximum bounds of the manipulable Cartesian workspace $\mathcal{T}_m \subset \mathcal{T}$. This allows us to differentiate between the reachable region of the manipulator workspace and the manipulable region of the workspace as shown in Fig. \ref{figure:manip_filter}, where the manipulable region in front of the robot is defined with a radius interval $[r, R]$, height interval $[h, H]$, and an angle $\alpha$. The values of $[h, H]$ and $\alpha$ are chosen by plotting all valid configurations.
%$A^{-}$. %The radius interval is calculated as
%\begin{equation}
%    \begin{aligned}
%        r = \underset {i}{\text{min}} \, x_i, i \in 1..n \\
%        R = \underset {i}{\text{max}} \, x_i, i \in 1..n.
%    \end{aligned}
%\end{equation}
\\
\textcolor{Blue}{
With a robot located at the $i$'th pile segment, we define the segment's manipulability as the overlapping region between an arm's annulus sector and the mapped pile. Specifically, for each sample point in the annulus sector, we query the GPIS for its occupancy probability~\cite{kim-overlap} using its inferred signed distance $\mu$, variance $\sigma^2$ and normal $\mathbf{n}$. The sum of weighted occupancy for all samples defines the manipulability score:}
\textcolor{Blue}{
\begin{equation}
\Scale[0.90]{
\begin{aligned}
m[\mathbf{s}_i] =& \sum_{\mathbf{p}_j \in \mathcal{T}_m}  w_j \, p(o=1|\mathbf{p}_j) = \sum  {w_j \Phi(\frac{\alpha \mu_j+ \beta}{\sqrt{1+\alpha^2\sigma_i^2}})};  \\ 
\begin{bmatrix}\mu_j \\ \sigma_j \\ \mathbf{n}_j\end{bmatrix} =& f_{\text{IS}}(\mathbf{p}^{(\textrm{w})}_j), \quad w_j = \frac{\mathbf{n}_j \cdot \mathbf{p}^{(\textrm{w})}_j}{\|\mathbf{p}^{(\textrm{w})}_j\|}, \quad \mathbf{p}^{(\textrm{w})}_j = \mathbf{T}[\mathbf{s}_i] \, \mathbf{p}_j
\end{aligned}
}
\label{eqn:manip}
\end{equation}
}
\textcolor{Blue}{The more objects reachable by the robot in alignment with its orientation at the segment, the higher its $m$ score. 
Fig.~\ref{fig:utility}(\emph{b}) shows an example of the pose with highest $m$ score.}

\section{Evaluation}
\label{sec:evaluate}
We evaluated the performance of our framework with extensive simulated and real-life experiments. We built our mobile manipulator using Neobotix MP700 \cite{mp700} as the base and URe5 \cite{ur5e} for the arm \textcolor{Blue}{with a magnetic contact end-effector. An RGB-D camera is mounted on the mobile base. Our active mapping framework is developed according to the description in Section~\ref{sec:overview}.} It is written in C++/ROS and runs on a 6-core laptop.  Our test environment consists of piles of bricks on flat terrain. \textcolor{Blue}{All piles are roughly 4m$^2$ in size with $\sim$50 segments considered in the optimisation, which is trivially performed by finding the maximum amongst all candidate utility scores.} The bricks are labelled with AR-tags for easy pose detection with metal plates attached for magnetic grasping. This scenario resembles the necessary exploration and an object picking component in automatic construction tasks. A video accompanying our results can be found in https://tinyurl.com/y5qwn864.

\subsection{Simulation}
\begin{figure}[h]
	\centering
	\setlength\tabcolsep{3pt}	
	\begin{tabular}{ccc}	
\includegraphics[height=2cm,width=0.3\linewidth]{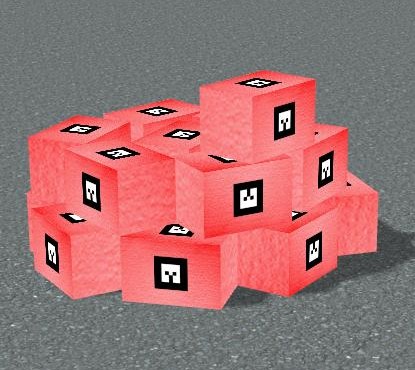} & \includegraphics[height=2cm,width=0.3\linewidth]{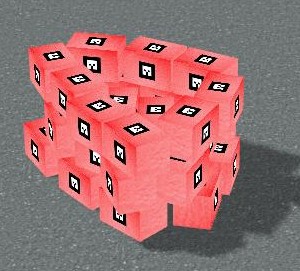} & \includegraphics[height=2cm,width=0.3\linewidth]{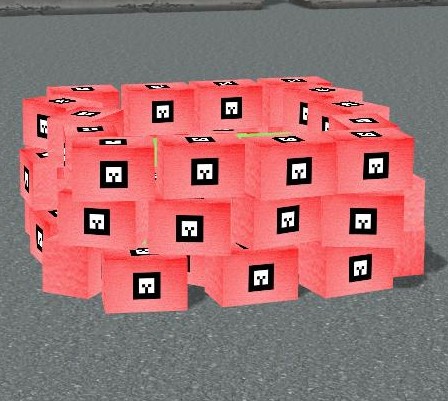}  \\
\small{(\emph{a}) simu case 1} & \small{(\emph{b}) simu case 2} & \small{(\emph{c}) simu case 3} 
\end{tabular}
\caption{\small{Three Gazebo simulation scenarios.}}
\label{fig:simu-scene}
\end{figure}

A simulated Gazebo environment was created to contain brick piles as shown in Fig.~\ref{fig:simu-scene}. Gazebo models of the base and arm were also designed with accurate mechanical properties to mimic the real-life system. An ablation study and benchmark tests are performed for the simulated environment. 

\subsubsection{Ablation Study}
We analysed the importance of each factor in our NBV selection scheme, by removing it from the framework and observing the performance results as given in Table~\ref{table:exp-ablation}. We run each variant in a fixed time interval and collect the average number of picked bricks, percentage of covered map, number of object falls and collisions between robot and environment. The ``Pick Order" row is important in maintaining a balanced pile shape, failing to do that the test has a high collapse count as seen in the ``Falls'' column -- a scenario to be avoided for warehousing or construction applications. The completion rate in ``Travel Distance" and ``Frontier" rows are related to collisions. The robot should never enter an unknown region before mapping it. Further, without the frontier factor, map coverage is affected; hence, achieving 100\% can take a protracted time. Without the ``Failure Penalty'' factor, the robot can get stuck indefinitely on a seemingly good spot that does not host pick-able objects. 
Considering all results, the complete utility function (row 1) case performs the best in each test category.
 
\begin{table}[th]
\caption{\small{Ablation analysis on Utility factors.}} 
\label{table:exp-ablation}
\centering
\footnotesize
	\setlength\tabcolsep{3pt}	
\begin{tabular}{||l||c|c|c|c||}
\hline
Part removed & \multicolumn{4}{c||}{Output } \\
\cline{2-5} 
 & Bricks \# & Map \% & Falls \# & Collision ?\\
 %& \# & \% & \# & Y/N\\
 \hline
NONE & \textbf{17} & \textbf{100} & \textbf{1} & \textbf{N}   \\
\hline
Manipulability &  14.5 & 100 & 2 & N  \\\hline
Pick Order & 13 & 100 & 6 & N \\\hline
Travel Distance & 11 & 80 & 3 & Y  \\\hline
 Uncertainty & 15 & 100 & 1 & N\\\hline
 Frontier & 16 & 90 & 4 & Y \\\hline
 Failure penalty & 7 & 60 & 1 & N \\\hline
 \end{tabular}
\end{table}

\subsubsection{Benchmark Test}
\begin{figure}[h]
\centering
	\begin{tabular}{cc}	
\tcbox[top=0pt,left=1pt,right=1pt,bottom=0pt, colback=white,colframe=blue!70!green]{\includegraphics[clip, trim=0.5cm 0.1cm 1.9cm 0.2cm,height=3.4cm]{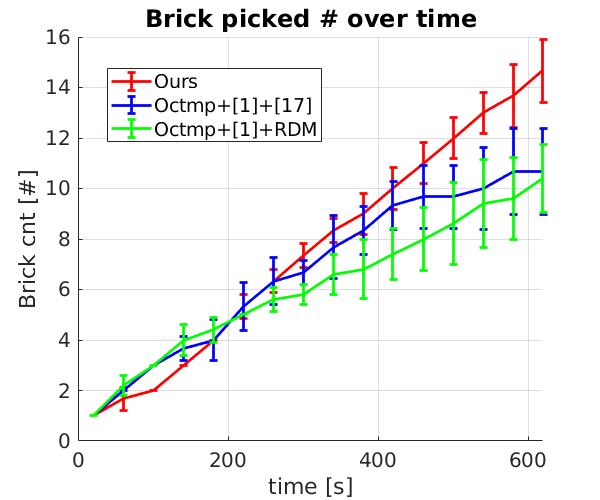}} &
\tcbox[top=0pt,left=1pt,right=1pt,bottom=0pt, colback=white,colframe=blue!70!green]{\includegraphics[clip, trim=0.5cm 0.1cm 1.9cm 0.2cm,height=3.4cm]{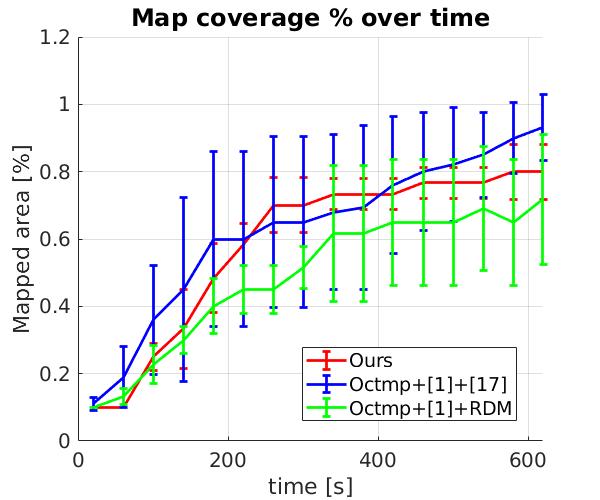}} \\
(\emph{a}) Objects picked &
(\emph{b}) Map coverage \% \\
\end{tabular}
\caption{\small{Comparison of task and map coverage for simu case 2.}}
\label{fig:exporation-test}
{\vspace{-5pt}}	
\end{figure}

We performed benchmark testing by \textcolor{Blue}{comparing task throughput and map coverage between our system and two other systems for three test scenarios, five times each (Fig.~\ref{fig:simu-scene})}. \textcolor{Blue}{The other systems are: \emph{(1)} Octomap + \cite{simon-acra} (dynamic) + \cite{entropy-gradient}  (discrete gradient frontier), and \emph{(2)} Octomap + \cite{simon-acra} (dynamic) + random (RDM) strategy.} The results are shown in Table \ref{table:benchmark}, showing the number of bricks picked (task) and map coverage. \textcolor{Blue}{From the results, our framework outperforms \emph{(1)} and \emph{(2)} in mission completion rate.} We observed in \emph{(1)} and \emph{(2)} that the robot frequently went to locations that do not host pick-able objects. \textcolor{Blue}{This can be explained by the fact that our framework uses a GPIS-based manipulability factor that chooses more promising locations for picking.} \textcolor{Blue}{\emph{(1)} has the best map coverage but can occasionally lead the robot into the obstacle zone, this can be explained by its approximate frontier calculation. (\emph{2}) has the lowest map coverage as it tends to stay in already explored regions.}
\textcolor{Blue}{We also illustrate the execution progress of the three systems for test case 2 in Fig.~\ref{fig:exporation-test}. Our approach has the best performance for the picking object task and equivalent performance for map coverage, compared to Octomap +[1]+[17], but with smaller variations over the different runs.
}
%our approach has the best performance for the picking object task and equivalent performance for map coverage than Octomap +[1]+[17], but with smaller variations over the different runs.
\begin{table}[hb]
{\vspace{-10pt}}	\caption{\small{Benchmark Test}} 
\label{table:benchmark}
\centering
\setlength\tabcolsep{1pt}
\begin{tabular}{||l||c|c||c|c||c|c||}
\hline
Simu & \multicolumn{2}{c||} {\footnotesize{Ours}} & \multicolumn{2}{c||}{ \textcolor{Blue}{
\footnotesize{\emph{(1)}  Octmp+\cite{simon-acra}+\cite{entropy-gradient}}}} 
& \multicolumn{2}{c||}{ \textcolor{Blue}{\footnotesize{\emph{(2)} Octmp+\cite{simon-acra}+RDM} }} \\ 
\cline{2-7}
 case & \footnotesize{\emph{Bricks\%}} & \footnotesize{\emph{Map\%}} & \footnotesize{\emph{Bricks\%}} & \footnotesize{\emph{Map\%}} & \footnotesize{\emph{Bricks\%}} & \footnotesize{\emph{Map\%}}\\
\cline{2-7}
\hline \hline 
\#1   &  
\textcolor{Blue}{\footnotesize{\textbf{95}$\pm$5.0}} &  \textcolor{Blue}{\footnotesize{\textbf{100}$\pm$10}}  & \textcolor{Blue}{\footnotesize{63$\pm$17}}  & \textcolor{Blue}{\footnotesize{\textbf{100}$\pm$5.0}} &
\textcolor{Blue}{\footnotesize{45$\pm$12}} & \textcolor{Blue}{\footnotesize{90$\pm$10}}\\
\hline
\#2  & 
\textcolor{Blue}{\footnotesize{\textbf{100}$\pm$6.0}} &  \textcolor{Blue}{\footnotesize{82$\pm$8.5}} &  \textcolor{Blue}{\footnotesize{72$\pm$8.0}} &  \textcolor{Blue}{\footnotesize{\textbf{85}$\pm$9.0}} & \textcolor{Blue}{\footnotesize{61$\pm$12}} & \textcolor{Blue}{\footnotesize{71$\pm$19}}  \\
\hline
\#3  & \textcolor{Blue}{\footnotesize{\textbf{100}$\pm$3.0}} & 
\textcolor{Blue}{\footnotesize{61$\pm$3.5}} & \textcolor{Blue}{\footnotesize{66$\pm$13}} &  \textcolor{Blue}{\footnotesize{\textbf{70}$\pm$16}} & \textcolor{Blue}{\footnotesize{57$\pm$6.8}} & \textcolor{Blue}{\footnotesize{57$\pm$12}}\\ 
\hline
\end{tabular}
{\vspace{-10pt}}	
\end{table}

\subsection{Real-life Experiment}
\begin{comment}
\begin{figure}[h]
\centering
	\begin{tabular}{cc}	
\setlength\tabcolsep{1pt}
\includegraphics[height=3cm,width=0.4\linewidth]{figures/real_life/scene1/scene-1_setup_small.jpg}
&
\includegraphics[height=3cm,width=0.4\linewidth]{figures/real_life/scene3/sene_3.jpg} \\
\small{(\emph{a}) Scene 1} & \small{(\emph{b}) Scene 2}
\end{tabular}
\caption{\small{Real-life experiment setup with our mobile manipulator in front of a pile of bricks.}}
\label{fig:reallife-setup}
{\vspace{-5pt}}	
\end{figure}
\end{comment}
In the real-life experiment, we tested the accuracy of the dynamic GPIS mapping component and the effectiveness of NBV selection in our framework. 

\subsubsection{Dynamic GPIS accuracy} 
\begin{figure}[h]
	\centering
	\setlength\tabcolsep{0pt}	
	\begin{tabular}{llllll}
\multicolumn{2}{l}{
\includegraphics[height=2.5cm,width=0.33\linewidth]{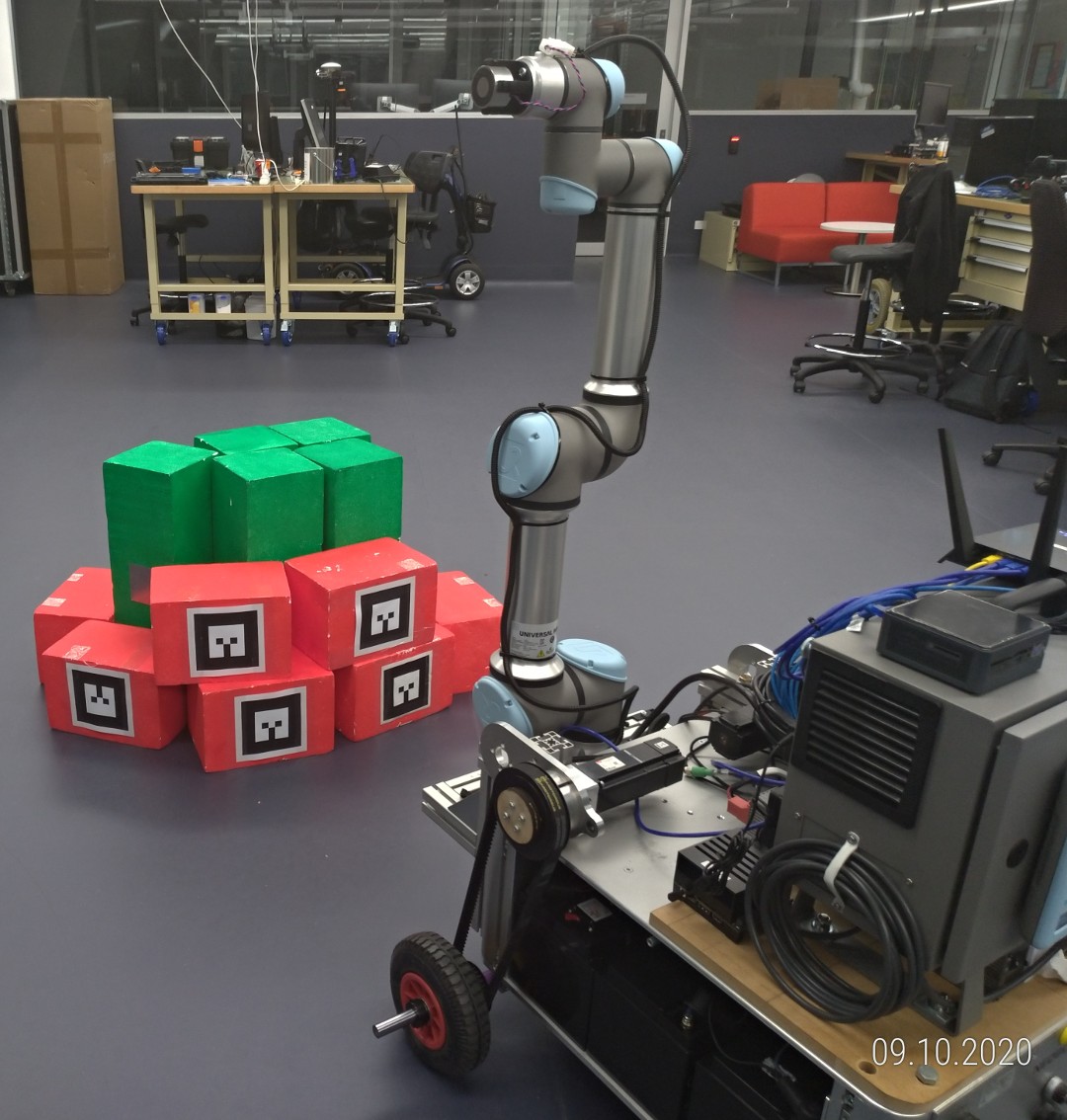}
}
&
\multicolumn{2}{l}
{
%    \def\stackalignment{r}
%    \topinset{
%\tcbox[arc=1pt,boxrule=0.5pt,top=-2pt,left=-2pt,right=-2pt,bottom=-2pt,colback=gray,colframe=gray]{    \includegraphics[height=0.7cm,width=0.8cm]{./figures/real_life/scene1/rgb_before_large.jpg}}}
%    {\tcbox[arc=1pt,top=-1pt,left=-1pt,right=-1pt,bottom=-1pt,colback=white,outer arc=0pt, colframe=gray]{	\includegraphics[height=2.25cm,width=0.3\linewidth]{./figures/real_life/scene1/persp2/before_2-modified.jpg}}}
%    {1pt}{1pt}
    \def\stackalignment{l}
    \topinset{
\tcbox[arc=1pt,boxrule=0.5pt,top=-2pt,left=-5pt,right=-2pt,bottom=-2pt,colback=gray,colframe=blue]{    \includegraphics[height=0.8cm,width=1.2cm]{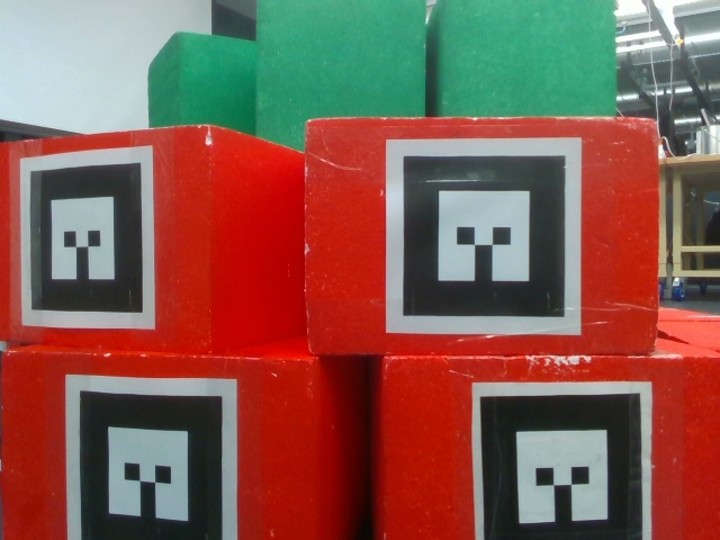}}}
    {\tcbox[arc=1pt,top=-1pt,left=-3pt,right=-1pt,bottom=-3pt,colback=white,outer arc=0pt, colframe=gray]{	\includegraphics[height=2.35cm,width=0.31\linewidth]{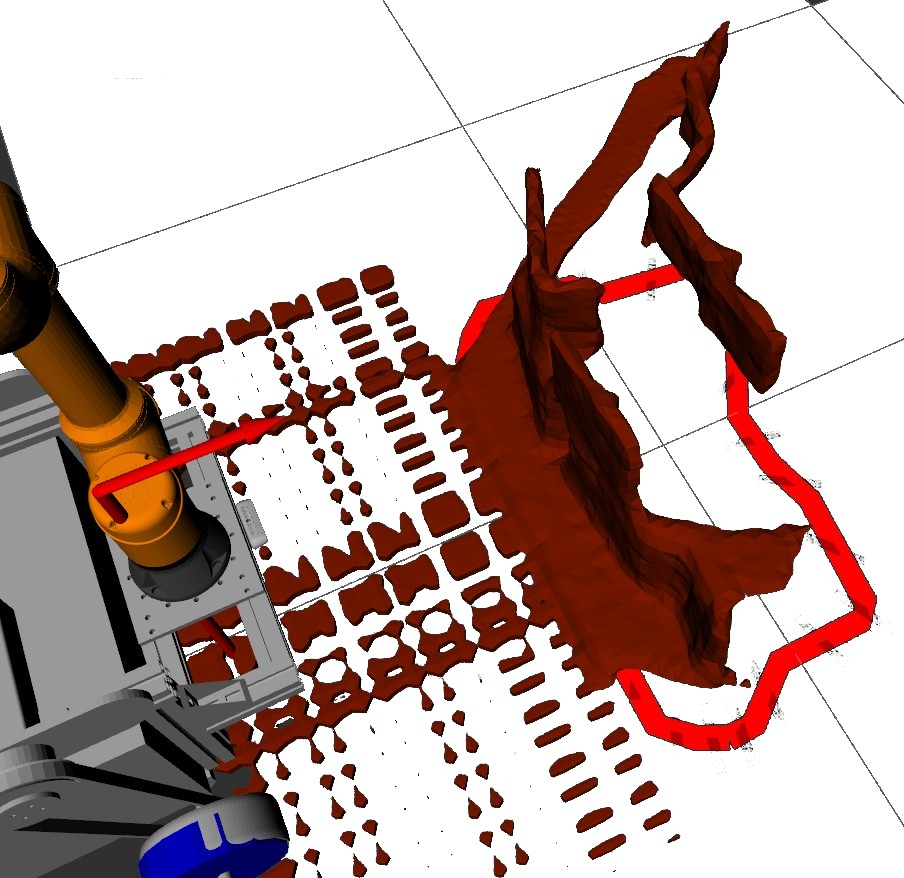}}}
    {0pt}{0pt}
}
&
\multicolumn{2}{l} {
%    \def\stackalignment{r}
%    \topinset{
%\tcbox[arc=1pt,boxrule=0.5pt,top=-2pt,left=-2pt,right=-2pt,bottom=-2pt,colback=gray,colframe=gray]{    \includegraphics[height=0.7cm,width=0.8cm]{./figures/real_life/scene1/rgb_after_large.jpg}}}{
%    \tcbox[arc=1pt,top=-1pt,left=-1pt,right=-1pt,bottom=-1pt,colback=white,outer arc=0pt, colframe=gray]{	\includegraphics[height=2.25cm,width=0.28\linewidth]{./figures/real_life/scene1/persp2/after-modified.jpg}}}
%    {1pt}{1pt}
    \def\stackalignment{l}
    \topinset{
\tcbox[arc=1pt,boxrule=0.5pt,top=-2pt,left=-2pt,right=-2pt,bottom=-2pt,colback=gray,colframe=blue]{    \includegraphics[height=0.8cm,width=1.2cm]{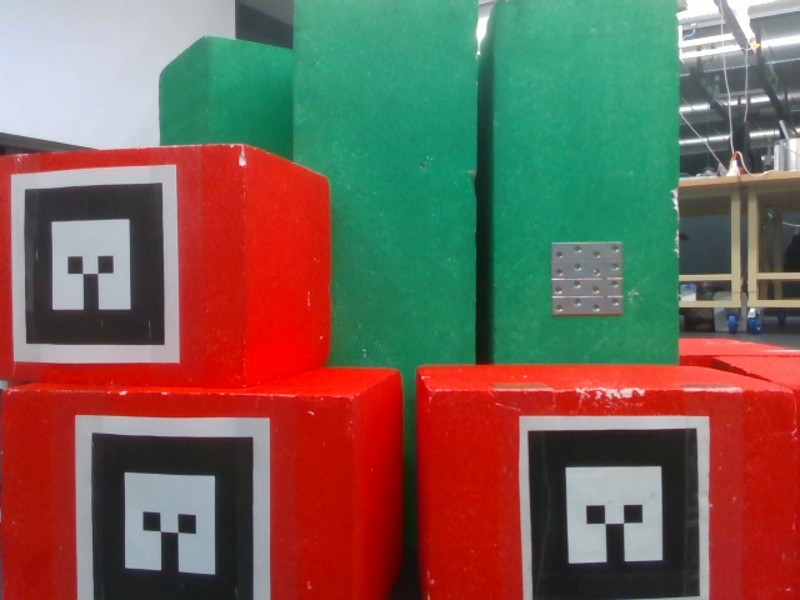}}}{
    \tcbox[arc=1pt,top=-1pt,left=-4pt,right=-1pt,bottom=-3pt,colback=white,outer arc=0pt, colframe=gray]{	\includegraphics[height=2.35cm,width=0.33\linewidth]{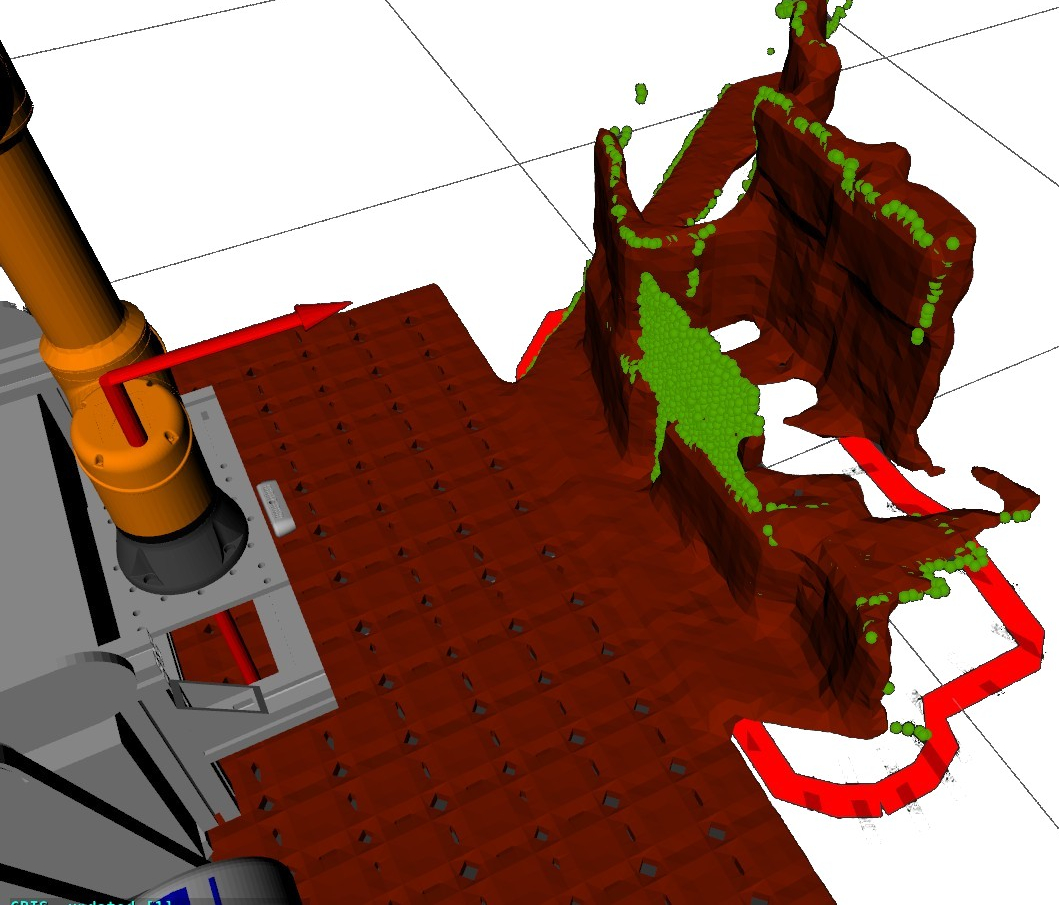}}}
    {0pt}{0pt}
}
\\ 
\multicolumn{2}{c} {
	\small{\begin{tabular}{@{}l@{}}(\emph{a})    Scene set up\end{tabular}} 
}	
&
\multicolumn{2}{c} {
	\small{\begin{tabular}{@{}l@{}}(\emph{b})    Initial GPIS \end{tabular}} 
}	
&
\multicolumn{2}{c}{
\footnotesize{\begin{tabular}{@{}l@{}}(\emph{c}) Remove brick, delete\\GPIS samples (green)
         \end{tabular}} 
}
\\
\multicolumn{3}{c}{
	\includegraphics[height=2.5cm,width=0.45\linewidth]{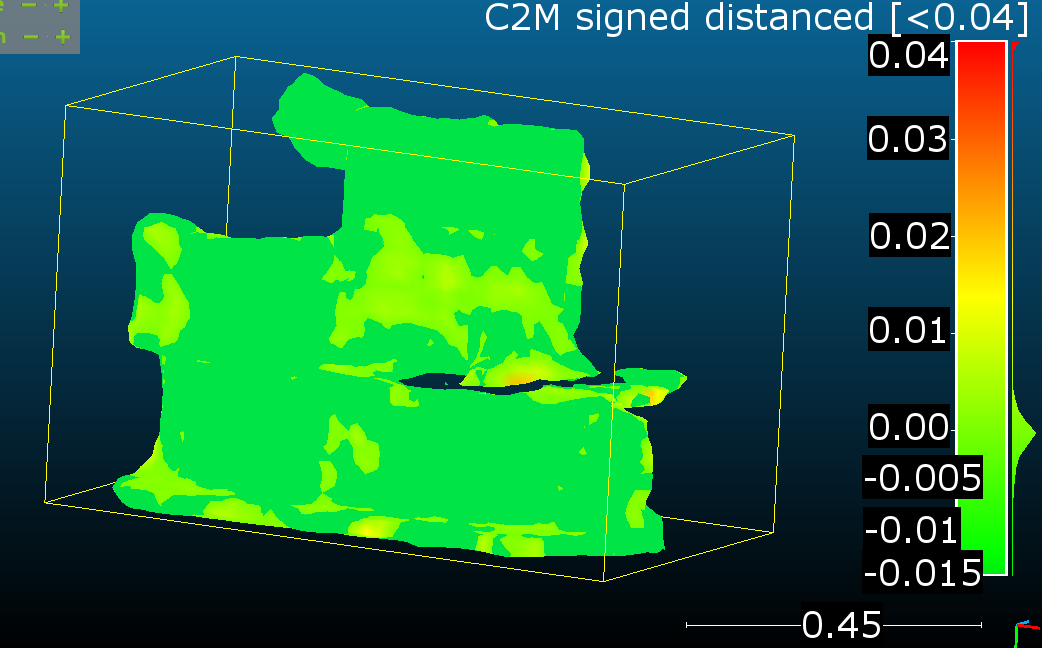}
}
&
\multicolumn{3}{c}{
	%\tcbox[top=0pt,left=2pt,right=2pt,bottom=0pt, colback=white,colframe=blue!70!green]
	{\includegraphics[height=2.5cm, width=0.45\linewidth]{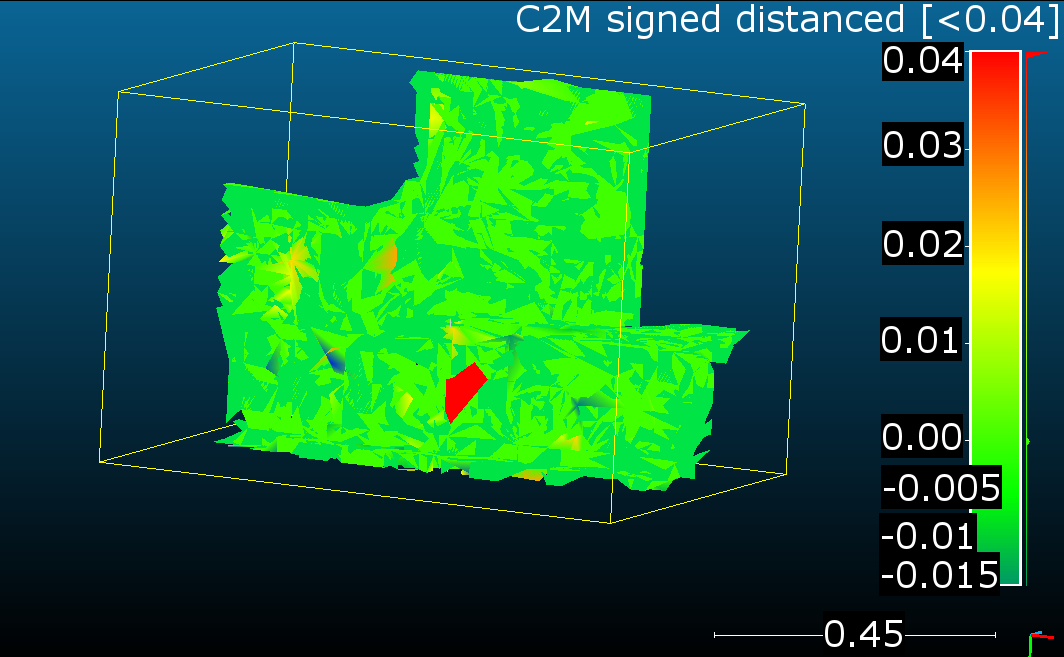}}
}
\\

\multicolumn{3}{c}{\textcolor{Blue}{
	\small{\begin{tabular}{@{}l@{}}(\emph{d})    Dynamic GPIS map vs \cite{bhoram-lee},\\ (note the smooth surface), \\obtain distance errors: \\
	$\Scale[0.85]{\mu_{dist}=0.005}$ cm, $\Scale[0.85]{\sigma_{dist}=0.012}$\end{tabular}}} 
}	
		& 
\multicolumn{3}{c}{	\small{\textcolor{Blue}{\begin{tabular}{@{}l@{}}(\emph{e})    Apply \cite{simon-acra}, obtain before \\and after Octomaps, (note the \\artifacts), obtain distance errors: \\ 
	$\Scale[0.85]{\mu_{dist}=0.033}$ cm, $\Scale[0.85]{\sigma_{dist}=0.015}$\end{tabular}}} 
}
\\
	\end{tabular}			
	\caption{\small{Real-life experiment: comparing map accuracy between ours, \cite{bhoram-lee} and \cite{simon-acra}}.}
{\vspace{-10pt}}	
	\label{fig:scene1}
\end{figure}
We set up a real-life scene Fig.~\ref{fig:scene1}(\emph{a}) to evaluate the accuracy of dynamic GPIS \textcolor{Blue}{by comparing with the methods of \cite{bhoram-lee} and \cite{simon-acra}}. We first generated a map using our Dynamic GPIS with a 2-step process: initialise using the original scene (\emph{b}), then update after removing a brick (\emph{c}). \textcolor{Blue}{For reference, we generated a map by feeding the depth image in step 2 directly to \cite{bhoram-lee}}. We compared the signed distance values in the two maps using CloudCompare (\emph{d}), and the error was insignificant (\emph{e}). \textcolor{Blue}{Further, we applied the same procedure for the Octomap variant \cite{simon-acra} and our results are shown to be superior (\emph{e})}.

\subsubsection{NBV test}
~\label{fig::scene2-exp}
	 \begin{figure*}[th]
	\centering
	{\vspace{-5pt}}	
	\setlength\tabcolsep{1pt}	
    \begin{tabular}{ccccc}
\includegraphics[height=2.5cm,width=0.19\linewidth]{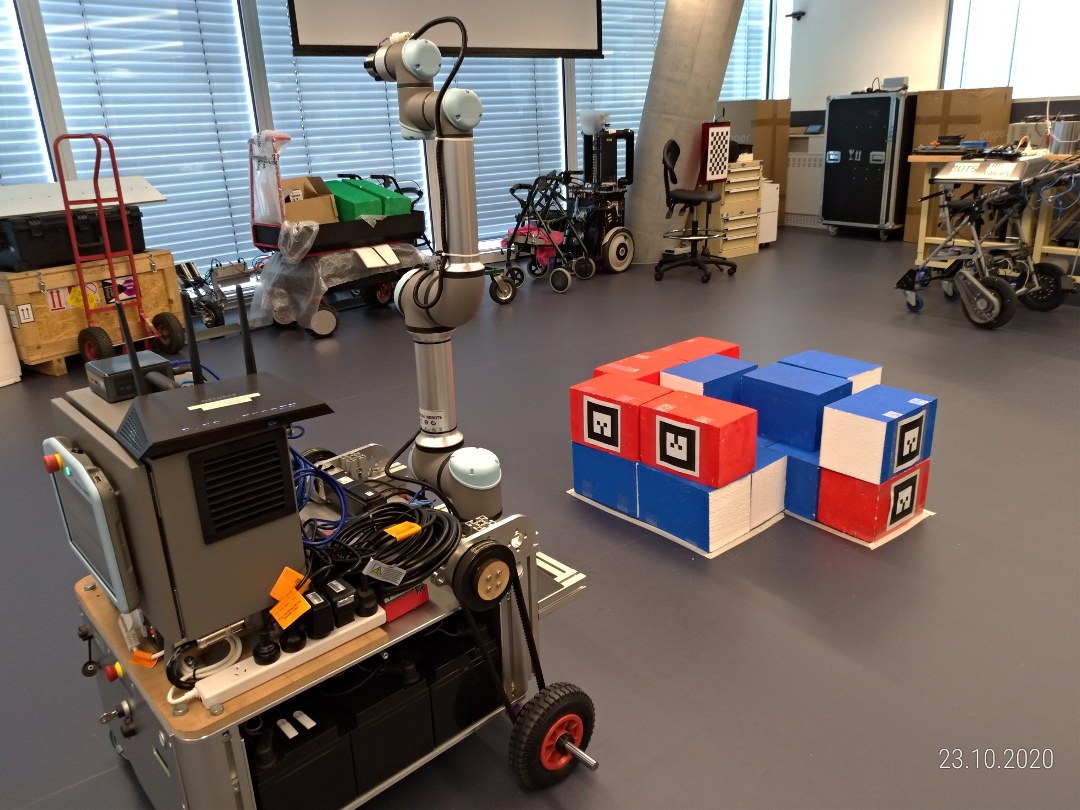}
&
    \def\stackalignment{r}
    \topinset{
\tcbox[arc=0pt,boxrule=0.5pt,top=-2pt,left=-2pt,right=-2pt,bottom=-2pt,colback=gray,colframe=blue]{    \includegraphics[height=0.85cm,width=1.1cm]{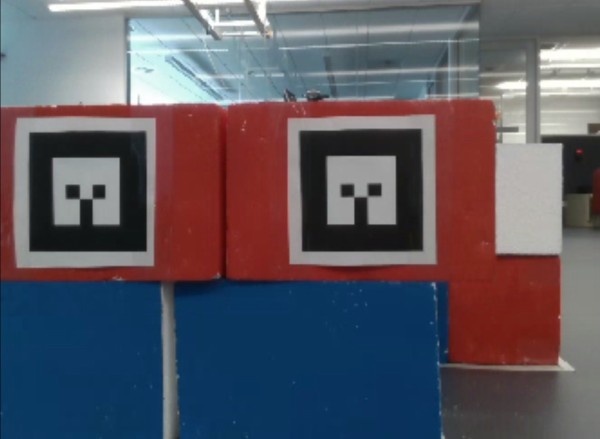}}}
    {\includegraphics[height=2.5cm,width=0.19\linewidth]{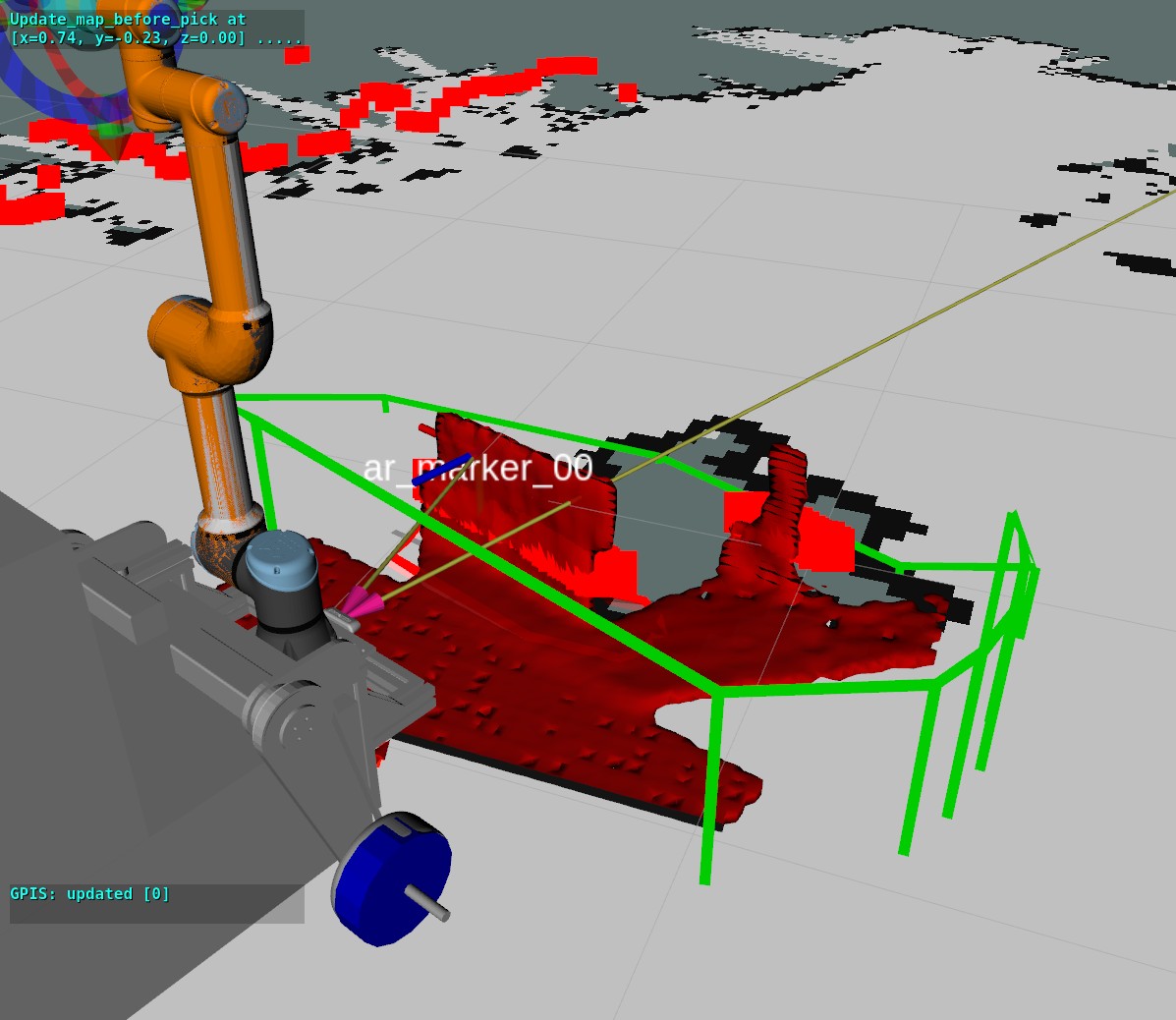}}
    {1pt}{0pt}
 &
     \def\stackalignment{r}
    \topinset{
\tcbox[arc=0pt,boxrule=0.5pt,top=-2pt,left=-2pt,right=-2pt,bottom=-2pt,colback=gray,colframe=blue]{     \includegraphics[height=0.85cm,width=1.1cm]{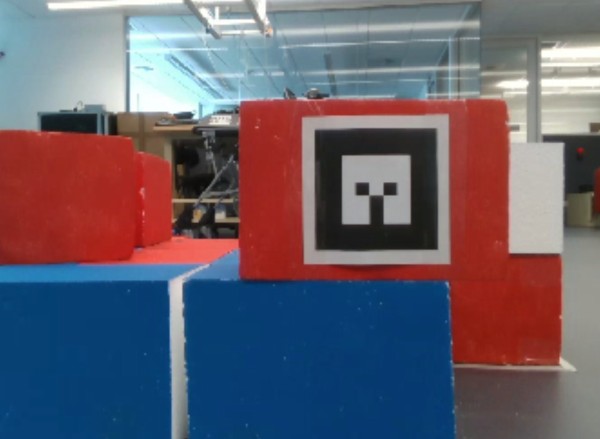}}}
    {\includegraphics[height=2.5cm,width=0.19\linewidth]{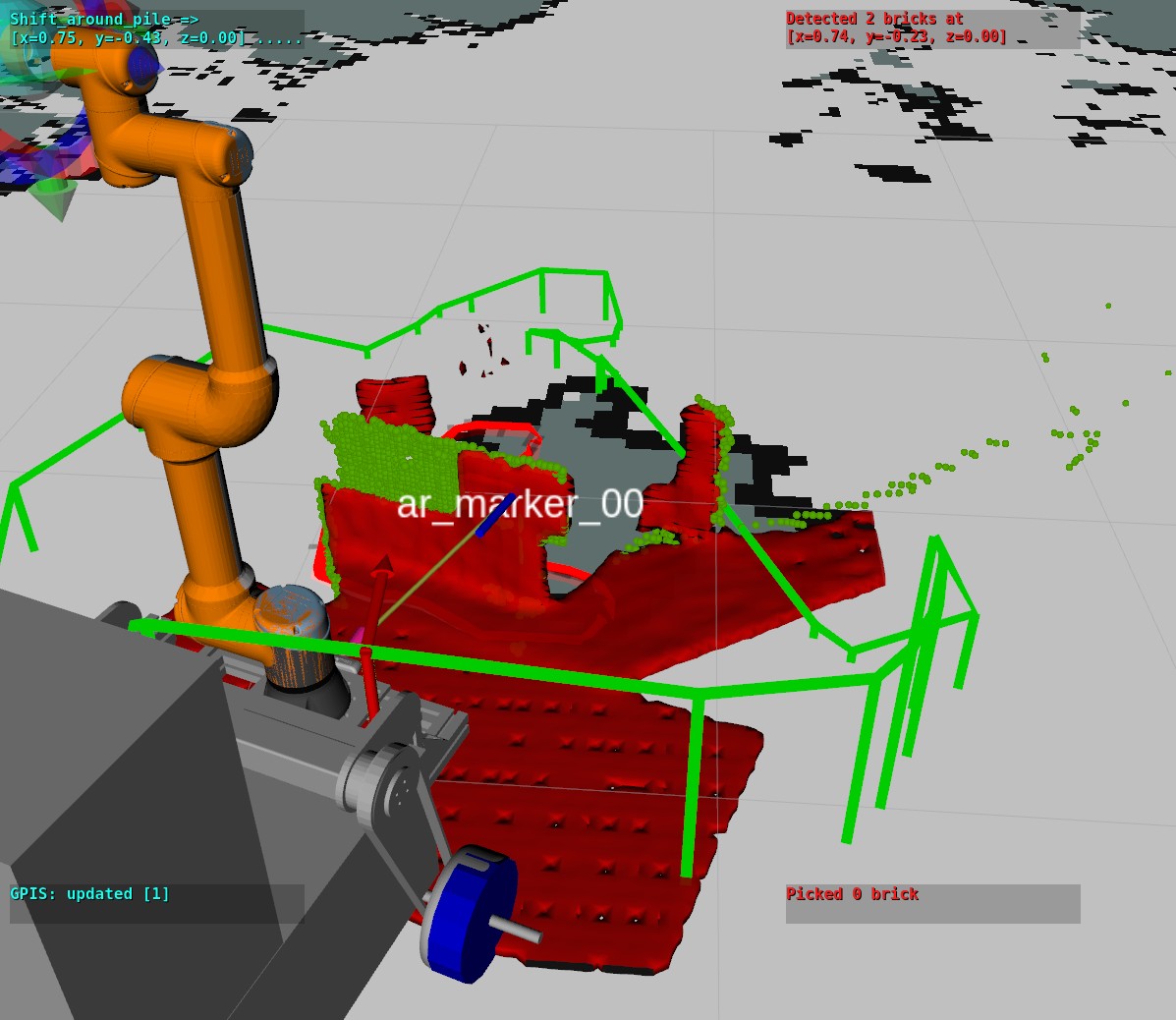}}
    {1pt}{0pt}
&
     \def\stackalignment{r}
    \topinset{
\tcbox[arc=0pt,boxrule=0.5pt, top=-2pt,left=-2pt,right=-2pt,bottom=-2pt,colback=gray,colframe=blue]{     \includegraphics[height=0.85cm,width=1.1cm]{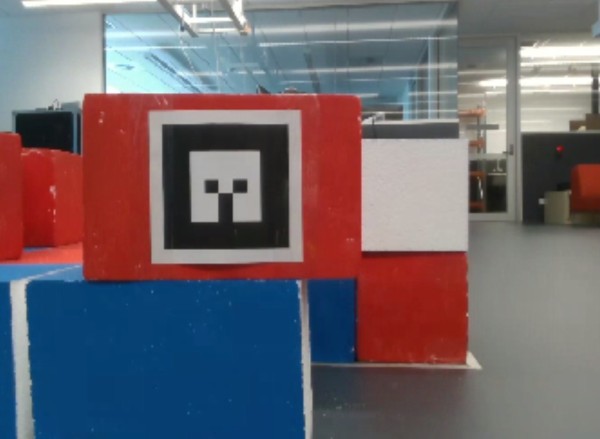}}}
    {\includegraphics[height=2.5cm,width=0.19\linewidth]{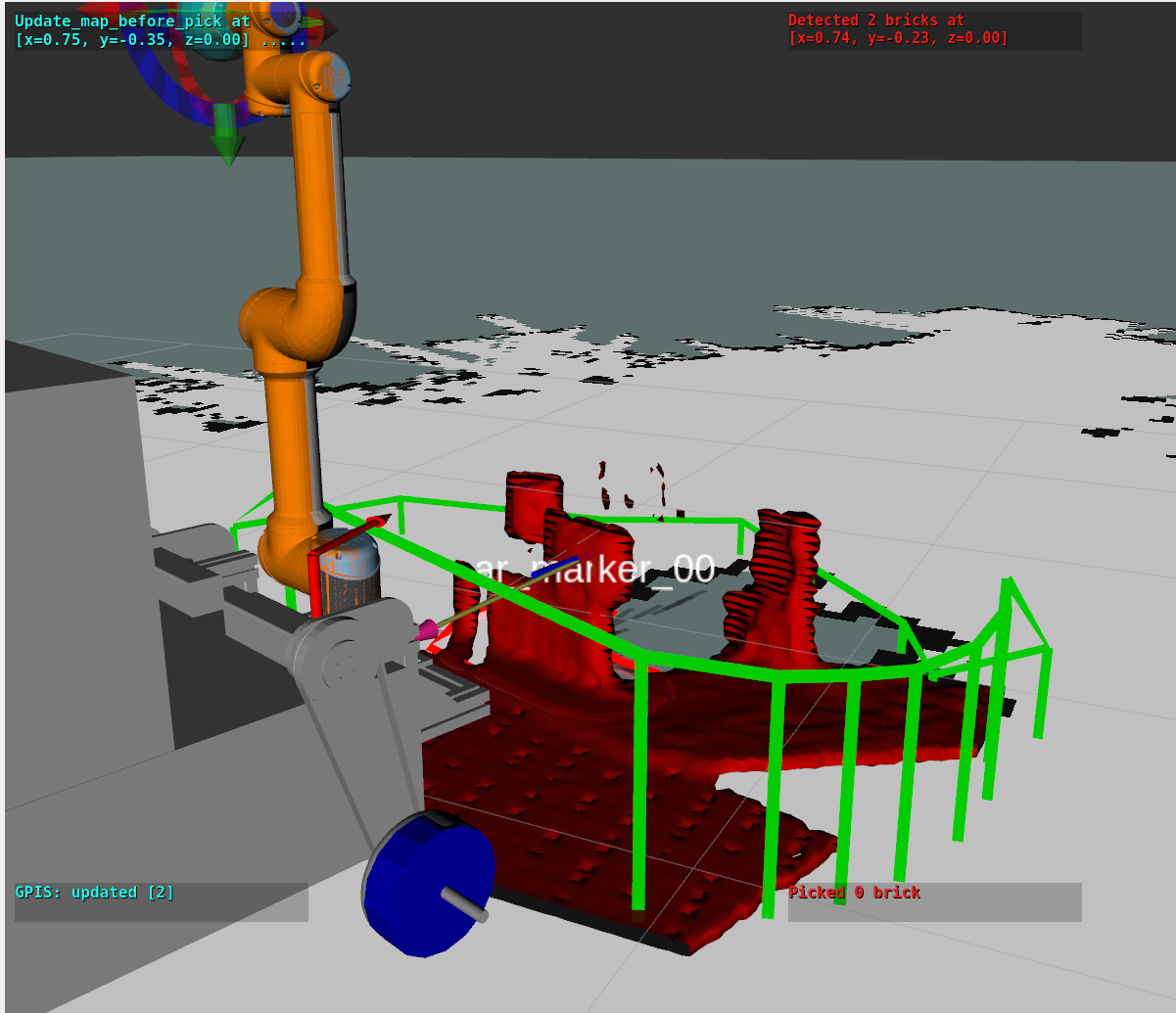}}
    {1pt}{0pt}
&
     \def\stackalignment{r}
    \topinset{
\tcbox[arc=0pt,boxrule=0.5pt,top=-2pt,left=-2pt,right=-2pt,bottom=-2pt,colback=gray,colframe=blue]{     \includegraphics[height=0.85cm,width=1.1cm]{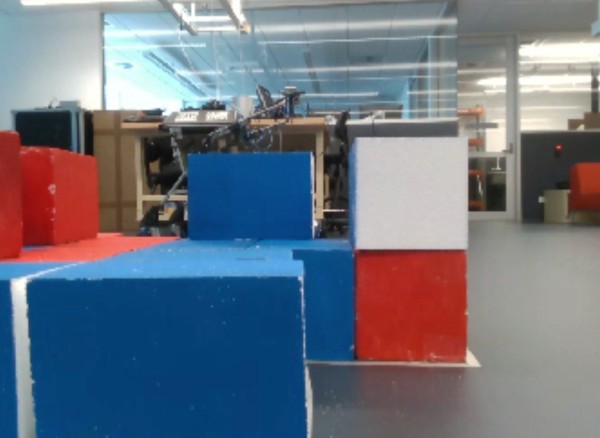}}}
    {\includegraphics[height=2.5cm,width=0.19\linewidth]{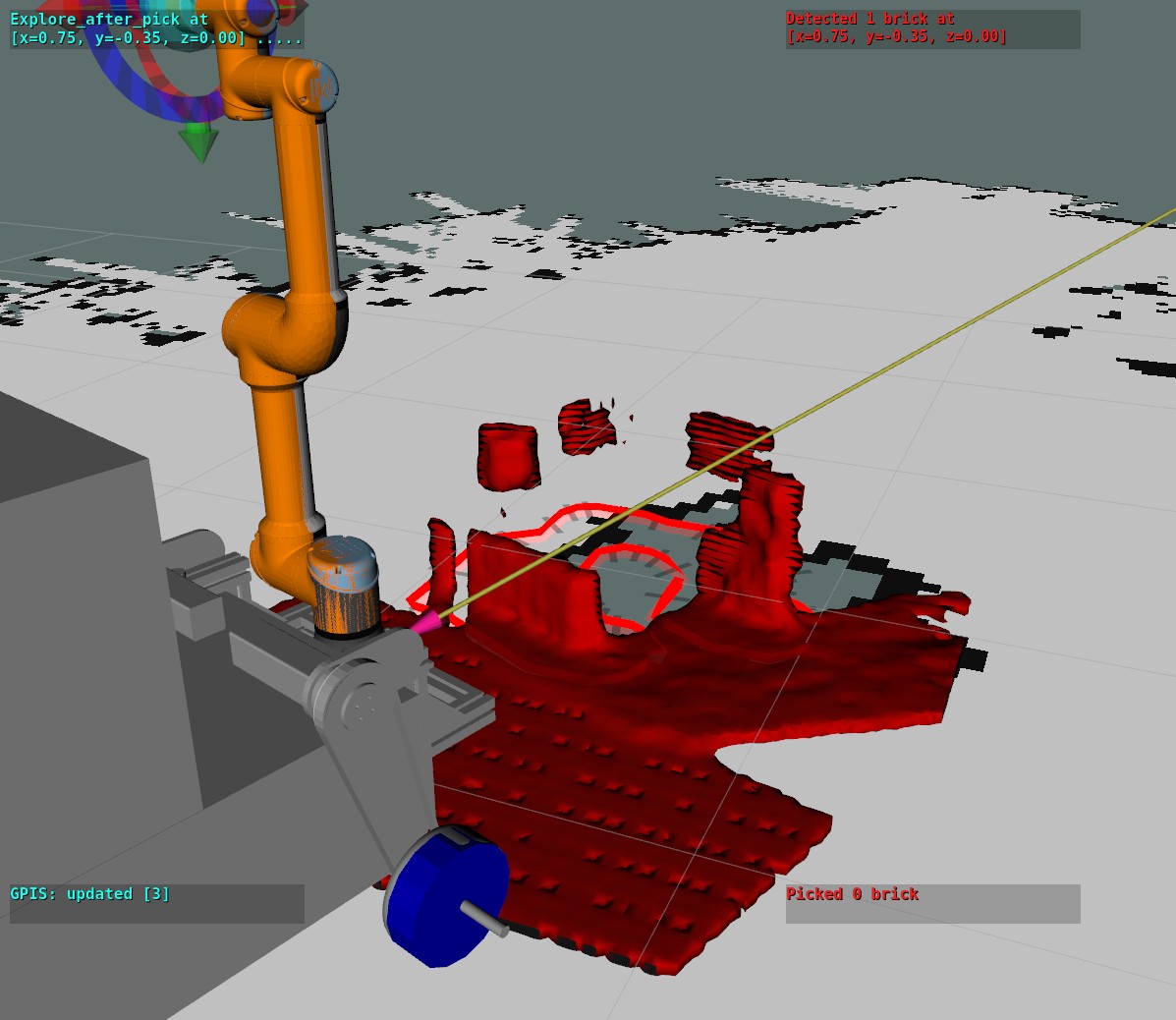}}
    {1pt}{0pt}
    \\
\small{(\emph{a}) test scene 2} &
\small{(\emph{b}) first scan} &
\small{(\emph{c}) picked one brick} & \small{(\emph{d}) moved to NBV} & 
\small{(\emph{e}) picked another brick}  \\
	\end{tabular}	
	{\vspace{-5pt}}
	\caption{ Real-life experiment, active and interactive mapping cycle.}
	{\vspace{-10pt}}
  	\label{fig::scene2-exp}
\end{figure*}
In real-life test scene 2 (Fig.~\ref{fig::scene2-exp}(\emph{a})), we evaluated the effectiveness of our NBV selection. We set the scene to have two rows of bricks with the top two lying side by side. Initially, the robot was set to face the top two bricks (\emph{b}). Then the robot detected and picked the top left brick, the map was updated and the NBV shifted towards the high brick on the right (\emph{c}). The robot then moved to the NBV position, (\emph{d}). Finally, it picks up the right brick. This shows our NBV strategy behaved in the desired order.

\section{Conclusion}
\label{sec::conclusion}
We presented an interactive and active mapping framework for a mobile manipulator platform based on dynamic GPIS. Since the framework is probabilistic, it is able to perform immediate mapping updates for a dynamically changing environment. Using the probabilistic map, our NBV selection scheme has been shown to balance the needs of information gain in visited regions, frontier driven map expansion, as well as object manipulability. Most importantly, the dense map generated enables a robot to safely move around in, and apply changes to, the environment. Both simulation and real-life experiments show our system can efficiently explore and interact with a large pile of objects in an environment.

%%%%%%%%%%%%%%%%%%%%%%%%%%%%%%%%%%%%%%%%%%%%%%%%%%%%%%%%%%%%%%%%%%%%%%%%%%%%%%%%

%%%%%%%%%%%%%%%%%%%%%%%%%%%%%%%%%%%%%%%%%%%%%%%%%%%%%%%%%%%%%%%%%%%%%%%%%%%%%%%%

%\bibliographystyle{IEEEtran}
%\bibliography{agp}

\begin{thebibliography}{10}
	\providecommand{\url}[1]{#1}
	\csname url@rmstyle\endcsname
	\providecommand{\newblock}{\relax}
	\providecommand{\bibinfo}[2]{#2}
	\providecommand\BIBentrySTDinterwordspacing{\spaceskip=0pt\relax}
	\providecommand\BIBentryALTinterwordstretchfactor{4}
	\providecommand\BIBentryALTinterwordspacing{\spaceskip=\fontdimen2\font plus
		\BIBentryALTinterwordstretchfactor\fontdimen3\font minus
		\fontdimen4\font\relax}
	\providecommand\BIBforeignlanguage[2]{{%
			\expandafter\ifx\csname l@#1\endcsname\relax
			\typeout{** WARNING: IEEEtran.bst: No hyphenation pattern has been}%
			\typeout{** loaded for the language `#1'. Using the pattern for}%
			\typeout{** the default language instead.}%
			\else
			\language=\csname l@#1\endcsname
			\fi
			#2}}
	
	\bibitem{simon-acra}
	S.~{Fryc}, L.~{Liu}, and T.~{Vidal-Calleja}, ``Efficient pipeline for mobile
	brick picking,'' in \emph{Australasian Conference on Robotics and Automation
		- (ACRA)}, 2019.
	
	\bibitem{xdbot}
	\BIBentryALTinterwordspacing
	IEEE, ``{Robotic Arm Wields UV Light Wand To Disinfect Public Spaces},'' 2020. [Online]. Available:
	\url{https://spectrum.ieee.org/news-from-around-ieee/the-institute/ieee-member-news/usc-researchers-robotic-arm-disinfect-coronavirus}
	\BIBentrySTDinterwordspacing
	
	\bibitem{warehouse}
	G.~Bartels and M.~Beetz, ``Perception-Guided Mobile Manipulation Robots for Automation of Warehouse Logistics,'' \emph{KI - Künstliche Intelligenz}, vol.~33, pp.189–192, 2019.
	
%	\bibitem{frontier-1997}
%	B.~{Yamauchi}, ``A frontier-based approach for autonomous exploration,'' in
%	\emph{Proceedings 1997 IEEE International Symposium on Computational
%		Intelligence in Robotics and Automation CIRA'97. 'Towards New Computational
%		Principles for Robotics and Automation'}, pp. 146--151.

	\bibitem{gp-book}
	C.~E. {Rasmussen} and C.~K.~I. {Williams}, ``Gaussian Processes for Machine Learning''.\hskip 1em plus 0.5em minus
	0.4em\relax MIT Press, 2006.
	
	
	\bibitem{masha-ipp}
	M.~{Popović}, T.~{Vidal-Calleja}, G.~{Hitz}, I.~{Sa}, R.~{Siegwart}, and
	J.~{Nieto}, ``Multiresolution mapping and informative path planning for
	UAV-based terrain monitoring,'' in \emph{2017 IEEE/RSJ International
		Conference on Intelligent Robots and Systems (IROS)}, pp. 1382--1388.
	
	\bibitem{Lauri2015ActiveOR}
	M.~Lauri, N.~Atanasov, G.~J. Pappas, and R.~Ritala, ``Active object recognition
	via monte carlo tree search,'' in \emph{International conference on robotics and automation (ICRA) Workshop }, 2015.
	
	\bibitem{fred-apple-picking}
	F.~{Sukkar}, G.~{Best}, C.~{Yoo}, and R.~{Fitch}, ``Multi-robot
	region-of-interest reconstruction with dec-mcts,'' in \emph{2019
		International Conference on Robotics and Automation (ICRA)}, pp.
	9101--9107.
	
	\bibitem{static-map-dynamic-object}
	C.~Jiang, D.~Paudel, Y.~Fougerolle, D.~Fofi, and C.~Demonceaux, ``Static-map
	and dynamic object reconstruction in outdoor scenes using 3-d motion
	segmentation,'' \emph{IEEE Robotics and Automation Letters}, vol.~1, pp.
	1--1, 01 2016.
	
	\bibitem{DynaSLAM}
	\BIBentryALTinterwordspacing
	B.~Bescos, J.~M. Facil, J.~Civera, and J.~Neira, ``Dynaslam: Tracking, mapping,
	and inpainting in dynamic scenes,'' \emph{IEEE Robotics and Automation
		Letters}, vol.~3, no.~4, p. 4076–4083, Oct 2018. 
	\BIBentrySTDinterwordspacing
	
	\bibitem{kinect-fusion}
	\BIBentryALTinterwordspacing
	R.~A. Newcombe, S.~Izadi, O.~Hilliges, D.~Kim, A.~J. Davison, P.~Kohli,
	J.~Shotton, S.~Hodges, and A.~Fitzgibbon, ``Kinectfusion: Real-time dense
	surface mapping and tracking,'' in \emph{IEEE ISMAR}.\hskip 1em plus 0.5em
	minus 0.4em\relax IEEE, October 2011.
	\BIBentrySTDinterwordspacing
	
	\bibitem{octomap}
	\BIBentryALTinterwordspacing
	A.~Hornung, K.~M. Wurm, M.~Bennewitz, C.~Stachniss, and W.~Burgard,
	``{OctoMap}: An efficient probabilistic {3D} mapping framework based on
	octrees,'' \emph{Autonomous Robots}, 2013, software available at
	\url{http://octomap.github.com}.
	\BIBentrySTDinterwordspacing
	
	\bibitem{GPIS}
	O.~Williams and A.~Fitzgibbon, ``Gaussian process implicit surfaces,'' 2007.
	
%	\bibitem{LanRAL20}
%	L.~Wu, R.~Falque, V.~Perez-Puchalt, L.~Liu, N.~Pietroni, and T.~Vidal-Calleja,
%	``Skeleton-based conditionally independent gaussian process implicit surfaces
%	for fusion in sparse to dense 3d reconstruction,'' \emph{2020 IEEE Robotics and
%		Automation Letters}, vol.~5, no.~2, pp. 1532--1539.

    \bibitem{kim-overlap}
  S.~Kim, J.~Kim. ``Continuous Occupancy Maps Using Overlapping Local Gaussian Processes,'' \emph{2013 IEEE/RSJ International Conference on Intelligent Robots and Systems}.

	\bibitem{bhoram-lee}
	B.~{Lee}, C.~{Zhang}, Z.~{Huang}, and D.~D. {Lee}, ``Online continuous mapping
	using gaussian process implicit surfaces,'' in \emph{2019 International
		Conference on Robotics and Automation (ICRA)}, pp. 6884--6890.
	
	\bibitem{gpis-shape}
	S.~{Dragiev}, M.~{Toussaint}, and M.~{Gienger}, ``Gaussian process implicit
	surfaces for shape estimation and grasping,'' in \emph{2011 IEEE
		International Conference on Robotics and Automation}, pp. 2845--2850.
	
	\bibitem{huang2019building}
	K.~Huang and T.~Hermans, ``Building 3d object models during manipulation by
	reconstruction-aware trajectory optimization,''
	\emph{ArXiv}, vol.~abs/1905.03907, 2019.
	
	\bibitem{entropy-gradient}
	R.~Rocha, J.~Dias and A.~Carvalho, ``Cooperative Multi-Robot Systems A study of Vision-based 3-D Mapping using Information Theory,'' \emph{2005 IEEE International Conference on Robotics and Automation},  pp.~384-389.
	
	\bibitem{maani-gp}
	M.~Jadidi, J.~Miro and G.~Dissanayake, ``Gaussian processes autonomous
	mapping and exploration for range-sensing mobile robots,'' \emph{Autonomous
		Robots}, no.~42, pp. 273--290, 2018.
	
	\bibitem{Yoshikawa85}
	T.~Yoshikawa, ``Manipulability of robotic mechanisms,'' \emph{The International
		Journal of Robotics Research}, vol.~4, no.~2, pp. 3--9, 1985.
	
	\bibitem{mp700}
	\BIBentryALTinterwordspacing
	N.~GmbH, ``Mobile robot mp-700,'' 2020. [Online]. Available:
	\url{https://www.neobotix-robots.com/products/mobile-robots/mobile-robot-mp-700}
	\BIBentrySTDinterwordspacing
	
	\bibitem{ur5e}
	\BIBentryALTinterwordspacing
	``{Universal Robot UR5e, a flexible and lightweight robotic arm},'' 2020.
	[Online]. Available:
	\url{https://www.universal-robots.com/products/ur5-robot}
	\BIBentrySTDinterwordspacing
	
	\bibitem{FSMI}
	  Z.~Zhang, T.~Henderson, S.~Karaman and V.~Sze, ``{FSMI: Fast computation of Shannon mutual information for information-theoretic mapping},'' \emph{The International Journal of Robotics Research}, vol.39, no.~9, pp. 1155–1177, 2020.

\bibitem{slam-occ}
F.~Bourgault, A.~Makarenko, S.~ Williams, B.~Grocholsky and H.~ Durrant-Whyte, ``Information based adaptive robotic exploration,'' \emph{ 2002 International Conference on Intelligent Robots and Systems}, pp. ~540-545 vol.~1.

\bibitem{active-grf-gpis}
S.~Caccamo, Y.~Bekiroglu, C.~H.~ Ek and D.~Kragic, ``Active exploration using Gaussian Random Fields and Gaussian Process Implicit Surfaces,'' \emph{2016 IEEE/RSJ International Conference on Intelligent Robots and Systems (IROS)}.


\bibitem{hilbert-dynamic}
R.~Senanayake, and F.~Ramos.   ``Bayesian Hilbert Maps for Dynamic Continuous Occupancy Mapping,'' \emph{Proceedings of the 1st Annual Conference on Robot Learning}, vol.~78, pp.~458-471.

%\bibitem{hilbert-realtime}
%V.~Guizilini, R.~Senanayake and F.~ Ramos, ``Dynamic Hilbert Maps: Real-Time Occupancy Predictions in Changing Environments,'' \emph{ 2019 International Conference on Robotics and Automation (ICRA)}.

\bibitem{ramos-2016}
F.~Ramos, L.~Ott. ``Hilbert maps: Scalable continuous occupancy mapping with stochastic gradient descent.'' \emph{The International Journal of Robotics Research}, 35(14), 2016.

\bibitem{first-dynamic}
T.~Hahnel, ``Map building with mobile robots in dynamic environments,'' \emph{2003 IEEE International Conference on Robotics and Automation}, vol. 2, pp. ~1557–1563.

\bibitem{marching-cube}
W.~Lorensen and H.~Cline. ``Marching cubes: A high resolution 3D surface construction algorithm,'' 1987 \emph{SIGGRAPH Comput. Graph}, vol.~21, pp.~163–169.

\bibitem{Gandler-explore}
G.~Gandler, C.~Ek, M.~Björkman, R.~Stolkin and Y.~Bekiroglu,
``Object shape estimation and modeling, based on sparse Gaussian process implicit surfaces, combining visual data and tactile exploration'',
\emph{Robotics and Autonomous Systems},
vol.~126, 2020.


\end{thebibliography}

\end{document}